\documentclass[10pt,journal,compsoc]{IEEEtran}

\ifCLASSOPTIONcompsoc
  \usepackage[nocompress]{cite}
\else
  \usepackage{cite}
\fi

\ifCLASSINFOpdf
\else
\fi

\usepackage{enumitem}
\usepackage{booktabs}
\usepackage{graphicx}
\usepackage{amsmath}
\usepackage{xcolor}
\usepackage{amssymb}
\usepackage{multirow}
\usepackage{subcaption}
\usepackage{tikz}
\usetikzlibrary{decorations.pathreplacing}
\usepackage[hidelinks]{hyperref}
\usepackage{array,mathtools}
\usepackage{soul}
\usepackage{enumitem}
\usepackage{algorithm}
\usepackage{algpseudocode}
\usepackage[mathscr]{euscript}
\usepackage{bbm}
\usepackage{xurl}

\usepackage[normalem]{ulem}
\useunder{\uline}{\ul}{}

\makeatletter
\let\OldStatex\Statex
\renewcommand{\Statex}[1][3]{%
  \setlength\@tempdima{\algorithmicindent}%
  \OldStatex\hskip\dimexpr#1\@tempdima\relax}
\makeatother

\DeclareMathAlphabet{\pazocal}{OMS}{zplm}{m}{n}
\newcommand{\unif}{\pazocal{U}}

\newcommand{\tikzmark}[2][]{\tikz[remember picture,overlay]\node[inner xsep=0pt,inner ysep=1ex,#1](#2){};}

\DeclareSymbolFont{rsfs}{U}{rsfs}{m}{n}
\DeclareSymbolFontAlphabet{\mathscrsfs}{rsfs} 

\newtheorem{definition}{Definition}[section]

\begin{document}

\title{Unsupervised Detection of  Behavioural Drifts with Dynamic Clustering and  Trajectory  Analysis}

\author{Bardh~Prenkaj ~and~Paola~Velardi 
\IEEEcompsocitemizethanks{\IEEEcompsocthanksitem Prenkaj and Velardi are with the Department of Computer Science, Sapienza University of Rome, Rome, Italy. \protect\\

E-mail: [surname]@di.uniroma1.it}
}

\markboth{}{Prenkaj and Velardi: Unsupervised Detection of  Behavioural Drifts with Dynamic Clustering and Trajectory Analysis}

\IEEEtitleabstractindextext{%
\begin{abstract}
Real-time monitoring of human behaviours, especially in e-Health applications, has been an active area of research in the past decades. On top of IoT-based sensing environments, anomaly detection algorithms have been proposed for the early detection of abnormalities. Gradual change procedures, commonly referred to as drift anomalies, have received much less attention in the literature because they represent a much more challenging scenario than sudden temporary changes (point anomalies). In this paper, we propose, for the first time, a fully unsupervised real-time drift detection algorithm named DynAmo, which can identify drift periods as they are happening. DynAmo comprises a dynamic clustering component to capture the overall trends of monitored behaviours and a trajectory generation component, which extracts features from the densest cluster centroids. Finally, we apply an ensemble of divergence tests on sliding reference and detection windows to detect drift periods in the behavioural sequence.
\end{abstract}

\begin{IEEEkeywords}
Anomaly detection, unsupervised detection, drift detection, behavioural changes, e-health, dynamic clustering.
\end{IEEEkeywords}}

\maketitle

\IEEEdisplaynontitleabstractindextext

\IEEEpeerreviewmaketitle

\section{Introduction}\label{sec:introduction}

Behavioural changes are gradual processes over a long period \cite{behaviouralchange}. Gradual change procedures represent a conceptually systematic set of behaviours \cite{behaviouralchange22},  widely analysed in many contexts; among these patterns of decline in the elderly resulting from Alzheimer's and Parkinson's diseases \cite{patternsofdecline}  and personal or collective behaviour changes, such as stopping smoking, saving energy and losing weight \cite{diet}. 

Recently, real-time monitoring systems based on sensors offer an unprecedented opportunity to monitor human behaviour \cite{sensordata} unobtrusively. For example, environmental sensors and wearable devices are widely used in telemedicine applications to support doctors in preventing, treating, and improving health conditions \cite{monitoring}. On top of these systems, deep learning anomaly detection algorithms have been proposed to automatically identify and detect behavioural changes, as surveyed in \cite{Cook2020AnomalyDF}. However, most models in the literature suffer from at least one of the following  limitations: 
\begin{enumerate}
\item They mostly concentrate on sudden, temporary changes, referred to as \textit{point anomalies} \cite{chandola2009anomaly}, rather than gradual changes (\textit{drift anomalies}).
\item They need training on behavioural data, which might not be realistic since behaviours, and anomalies therein, are highly context- and person-dependent.
\item They fail to discover latent drift periods when the training (reference) set contains anomalous behaviours, a possibility that cannot be ruled out in real-world contexts.
\end{enumerate}

This paper presents DynAmo, short for  \underline{\textbf{Dyn}}amic Drift \underline{\textbf{A}}no\underline{\textbf{m}}aly Detect\underline{\textbf{o}}r, a fully unsupervised strategy for detecting gradual behavioural changes based on dynamic clustering and trajectory detection. The dynamic clustering component captures an overall trend of the time series representing a monitored behaviour (e.g., sleeping). It produces clusters for each monitoring interval (e.g., one day). The densest cluster in each interval becomes the input to the next component. This trajectory generator extracts features from the cluster centroids. Finally, DynAmo predicts the drift areas for each observed feature of the monitored action, for example, the duration and onset of sleep or the number of sleep interruptions. Although the proposed strategy applies to general drift detection, this paper explicitly addresses a challenging scenario where the goal is unsupervised real-time detection of drift changes from sensor data sequences. This context is particularly relevant in telemedicine and continuous patient care \cite{surveymonitoringsystems}.

We organise the rest of this paper as follows. Sec. \ref{sec:related} discusses the related work and provides an overview of this paper's contribution. Sec. \ref{sec:methodology} describes our strategy, ranging from the input modelling techniques to the drift detection mechanism. Sec. \ref{sec:dataset} enlists synthetic and real-scenario datasets and illustrates their characteristics in the normal/anomalous period throughout the time series. Sec. \ref{sec:experiments} provides extensive experiments on DynAmo and SOTA methods. Finally, Sec. \ref{sec:conclusion} concludes the paper.

\section{Related Work}
\label{sec:related}

\begin{table}[!ht]
\centering
\caption{Drift anomaly detectors in the literature. In the Learning column, S denotes supervised learning, SS semi-supervised learning, U unsupervised learning.}
\label{tab:drift_detector_types}
\resizebox{\columnwidth}{!}{%
\begin{tabular}{lcccccc}
\toprule
\multirow{3}{*}{}                      & \multicolumn{4}{c}{Detector type}                      & \multirow{3}{*}{Learning} &
\multirow{3}{*}{Pub. year}\\
                                       & \multicolumn{2}{c}{Batch} & \multicolumn{2}{c}{Online} &                                 \\ \cmidrule{2-5}
 &
  \multicolumn{1}{l}{Whole-batch} &
  Partial-batch &
  \begin{tabular}[c]{@{}c@{}}Fixed reference\\ window\end{tabular} &
  \begin{tabular}[c]{@{}c@{}}Sliding reference\\ window\end{tabular} & 
   \\ \midrule
MD3 \cite{sethi2015don}                & \checkmark  & $\times$    & $\times$     & $\times$    & S          & 2015                     \\
IKS-bdd \cite{dos2016fast}             & $\times$    & $\times$    & \checkmark   & $\times$    & U          & 2016                     \\
SAND \cite{haque2016sand}              & $\times$    & $\times$    & $\times$     & \checkmark  & SS     & 2016                         \\
DbDDA \cite{kim2017efficient}          & $\times$    & $\times$    & $\times$     & \checkmark  & S    & 2016                    \\
CD-TDS \cite{koh2016cd}                & $\times$    & $\times$    & \checkmark   & $\times$    & U   & 2016                     \\
OMV-PHT \cite{lughofer2016recognizing} & $\times$    & $\times$    & \checkmark   & $\times$    & SS       & 2016          \\
UDetect \cite{bashir2017framework}     & \checkmark  & $\times$    & $\times$     & $\times$    & SS & 2017                    \\
MD3-EGM \cite{sethi2017reliable}       & \checkmark  & $\times$    & $\times$     & $\times$    & SS        & 2017    \\
DDAL \cite{costa2018drift}             & $\times$    & \checkmark  & $\times$     & $\times$    & S         & 2018               \\
NN-DVI \cite{liu2018accumulating}      & \checkmark  & $\times$    & $\times$     & $\times$    & U          & 2018              \\
PDetect \cite{sethi2018handling}       & $\times$    & \checkmark  & $\times$     & $\times$    & S      &      2018            \\
KLCPD \cite{DBLP:conf/iclr/ChangLYP19} & $\times$    & $\times$    & $\times$     & \checkmark  & U        & 2019                \\
D3 \cite{gozuaccik2019unsupervised} & $\times$ & $\times$    & $\times$     & \checkmark  & S       & 2019                \\
FAAD \cite{li2019faad}                 & \checkmark  & $\times$    & $\times$     & $\times$    & U   & 2019                     \\
Plover \cite{de2019learning}           & $\times$    & $\times$    & $\times$     & \checkmark  & U           & 2019             \\
DSDD \cite{pinage2020drift}            & $\times$    & $\times$    & $\times$     & \checkmark  & SS         & 2020               \\
ERICS \cite{haug2021learning} & $\times$    & $\times$  & $\times$     & \checkmark    & SS               & 2021         \\
STUDD \cite{cerqueira2022studd}       & \checkmark    & $\times$  & $\times$     & $\times$    & SS      & 2022                  \\ 
CDLEEDS \cite{DBLP:conf/cikm/HaugBZK22} & $\times$    & $\times$  & $\times$     & \checkmark    & U            & 2022            \\ 
 \midrule
 \textbf{DynAmo [us]}  & $\times$    & $\times$  & $\times$     & \checkmark    & U        & 2023                \\ 
 \bottomrule
\end{tabular}%
}
\end{table}

As we already remarked, while there is a vast literature on point anomaly detection \cite{Cook2020AnomalyDF}, drift detection received much less attention,  due to its increased complexity. 
As illustrated in Table \ref{tab:drift_detector_types}, systems of drift anomaly detection are divided into batch and online detectors. According to Gemaque et al. \cite{gemaque2020overview}, drift detection methods utilise a reference and detection window. The former (usually) contains the normal\footnote{We remark that, in real-life contexts, this is not guaranteed: anomalies may occur at any time, including the initial monitoring period.} event distribution, whereas the latter contains unseen data, possibly including sudden or drift anomalies. In \textit{batch detection} approaches, the reference window remains fixed in time, whereas the detection window slices through the trajectory of events. Batch detectors raise a drift anomaly when the distribution of the detection window differs from the reference window. Contrarily, the reference window of \textit{online detectors} is dynamically replaced by the detection window when their distributions differ more than an established threshold. This window change renders online detectors adjustable to routine changes (e.g. seasonality shifts, permanent or temporary changes of lifestyles). As the detection window moves through the series, the reference window can generally slide one event at a time\footnote{The amount of slide is a hyperparameter (e.g. every minute, every day, or every month).}.

The literature has contributed several \textit{(semi)supervised} solutions towards drift anomaly detection, whether by exploiting batch \cite{sethi2015don,bashir2017framework,sethi2017reliable,costa2018drift,sethi2018handling,cerqueira2022studd} or online approaches \cite{haque2016sand,kim2017efficient,lughofer2016recognizing,gozuaccik2019unsupervised,pinage2020drift,haug2021learning}. \textit{Unsupervised} drift detection has also been covered in the literature \cite{de2019learning,dos2016fast,koh2016cd,li2019faad,liu2018accumulating,DBLP:conf/cikm/HaugBZK22}.

\noindent\textbf{Batch drift detectors:} Liu et al. \cite{liu2018accumulating} propose NN-DVI. This distribution-based approach assumes that regional density changes cause miss-drifts. The authors rely on three modules: (1) kNN-based space partitioning for data modelling, (2) distance function to accumulate density discrepancies, and (3) statistical significance test to determine the drifts. Li et al. \cite{li2019faad} build models using random feature sampling and calculate their corresponding anomaly scores. They exploit an anomaly buffer based on a model dynamic adjustment algorithm to distinguish between true drifts and normal sequences incorrectly labelled anomalous. Bashir et al. \cite{bashir2017framework} propose a two-phase architecture. First, they train a classifier and collect data characterising each class. Second, they collect batches of data for each class and verify whether the instances of these classes differ from the data of the previous phase. Sethi and Kantardzic \cite{sethi2015don} propose MD3 to monitor changes in the region of the classifier decision space where predictions are uncertain. The authors assume that drift occurs when the density of this variation is higher than a specific threshold, similar to \cite{costa2018drift}. Inspired by the limitation of \cite{sethi2015don}, the same authors \cite{sethi2017reliable} propose MD3-EGM, a semi-supervised method based on ensembles. Lastly, Cerqueira et al. \cite{cerqueira2022studd} propose STUDD. In this semi-supervised teacher-student learning paradigm, drifts are detected according to the error of the student model.

\noindent\textbf{Online drift detectors:} dos Reis et al. \cite{dos2016fast} propose IKS-bdd, an online form of the KS test with two sliding windows for drift detection. Koh \cite{koh2016cd} proposes a drift detector on transactional data streams. The method has two parts: i.e. local and global drift detection. The main idea behind detecting local drifts is to compare W0 and W1 windows using the Hoeffding Bound. A drift is signalled when the sample mean difference between $W_0$ and $W_1$ is more than $\delta$. The author uses two decision trees for $W_0$ and $W_1$ for global drift detection and examines their disagreement. Lughofer et al. \cite{lughofer2016recognizing} use the Page-Hinkley test to detect changes. Chang et al. \cite{DBLP:conf/iclr/ChangLYP19} propose KLCPD, a composite kernel method that combines RBF kernels with injective functions. The authors parameterise the injective functions via RNNs to capture the temporal dynamics of complex time series. De Mello et al. \cite{de2019learning} exploit the concept of stability by computing the divergence between the sliding reference and detection windows. Haque et al. \cite{haque2016sand} propose SAND, an ensemble of kNN classifiers. SAND predicts the label of an unknown example $x^\prime$ by majority voting and stores the predicted class and confidence scores. If the confidence values diverge from the beta distribution by exceeding a threshold, then SAND detects a drift. An adaptation procedure occurs by updating the ensemble with the true labels of the instances with low confidence scores. Similarly, Pinagé et al. \cite{pinage2020drift} propose a dynamic classifier based on an initial ensemble and a configurable drift detector guided by a pseudo-error rate to perform detections. When a certain number of the base classifiers in the ensemble indicate a drift, the validation set gets updated using the new labelled samples; otherwise, the ensemble continues learning using the ground truth labels. Kim et al. \cite{kim2017efficient} first train the model on labelled instances. Then, they monitor the differences in uncertainty for the instances in both windows. Before labelling the instances of the detection window, the model calculates a confidence interval for the uncertainty of the events in the reference window. If it exceeds the upper limit, a drift is signalled. Upon detecting a drift, the authors retrain the model on the true positive events in the detection window. G{\"o}z{\"u}a{\c{c}}{\i}k et al. \cite{gozuaccik2019unsupervised} introduce D3. It integrates a discriminative classifier in an online fashion. By analysing the classifier's performance on a fixed-size sliding window, D3 identifies concept drifts without estimating distributions. AUC (Area Under the Curve) measures separability between old and new data classes. When AUC falls below a threshold, indicating insufficient separability, drifts are detected. The sliding window is adjusted dynamically by removing or replacing samples based on the underlying classifier's performance. Haug and Kasneci \cite{haug2021learning} propose ERICS, a model-agnostic framework that treats the parameters of a predictive model as random variables and detects concept drifts by associating them with a change in the distribution of the optimal parameters. Lastly, to bridge the gap of invalid local attributions under drift conditions, Haug et al. \cite{DBLP:conf/cikm/HaugBZK22} propose CDLEEDS, an adaptive hierarchical clustering approach capable of detecting local and global distributional drifts.

\subsection{Limitations of the works in the literature and open challenges}
The works in the literature suffer from several limitations, including the cold-start problem, specifically for batch-based detectors. In detail, batch-based detectors do not cope with incoming real-time data. Hence, these detectors suffer in critical scenarios such as continual remote monitoring, which is the scope considered in this paper. Additionally, most of the methods enlisted in Table \ref{tab:drift_detector_types} require prior knowledge of the underlying distribution (data labels) to identify future drift periods correctly. Although knowing data labels is generally acceptable for non-impactful scenarios, \textit{one cannot assume to know human's normal behaviour beforehand}, especially for patients whose behaviour may be altered by their specific health conditions. For example, a disturbed sleep with many interruptions may be the normality for a certain patient, for whom, instead, a gradual change may be represented by the lengthening of the period in which they stay in bed.
Although unsupervised methods proposed in the literature overcome this problem,  they suffer from intra-window distributional changes and feature evolution. Therefore, while they might be able to identify a drift occurring inside the detection window correctly, \textit{they cannot detect a drift starting inside the reference window}. Since anomalies in real-world scenarios can occur anytime after the monitoring starts, not being able to detect them immediately is a drawback.
We argue that the outlined problem is a special case of the well-known cold-start problem since drifts can only be identified in the detection window. This phenomenon worsens if the detector is batch-based or an online detector with a fixed-reference window.

Another common limitation is that drift detectors are usually specialised in detecting only some kinds of drift types \cite{lu2018learning} (e.g., gradual or recurrent drifts). Thus, they are unsuitable to cope with sudden spikes of distributional changes since the width of the windows needs to be calibrated to handle the specific type of shift.

A final open issue is the reproducibility and replicability of the experiments provided in the original papers, which is essential to compare and evaluate the merits and drawbacks of the different solutions. While some works do not publish their code online, others do not thoroughly explain the data processing, hyperparameter tuning, and evaluation used\footnote{For example, do the methods employ soft margins to help detect drift periods? We refer to soft margins as the area before and after the drift - in terms of a particular time unit - that the prediction can still be considered correct/valid. As for gradual drifts, soft margins are typically used when a prediction at an exact time unit is unnecessary.}.

\subsection{Our contribution to the literature}
Considering the drawbacks of the SOTA described above, we provide  the following contributions:
\begin{enumerate}[nosep,topsep=0pt]
    \item We propose a \textit{fully unsupervised} drift detection technique based on dynamic clustering and trajectory detection, which works independently of input data distribution and prior knowledge of anomaly types (see Sec. \ref{sec:predictive_strategy} and Algorithm \ref{algo:dynamo}).
    
    \item We avoid the cold-start problem, frequently observed in the literature, by not reserving portions of the input to fine-tune the model to detect drifts.
    
    \item DynAmo is agnostic to various drift anomaly types (e.g., gradual and recurrent drift), which provides robustness w.r.t. other strategies in the literature.
    
    \item DynAmo has an integrated backward lookup parameter $\lambda$ that considers past events in a behavioural trajectory. DynAmo uses $\lambda$ to check the evolution of the feature hyperboxes associated with monitored events, which detects potential shifts within the same window (reference or detection) regardless of whether the distribution is anomalous (see Algorithm \ref{algo:detector} and Sec. \ref{sec:discussion}).
    
    \item DynAmo traces the trajectory of the densest cluster centroid for each sliding step, thus providing a visual and interpretable tool which gives domain non-specialists the ability to identify drifting trends in a two-dimensional space (see Sec. \ref{sec:dynamic_clustering} and \ref{sec:trajectory_generation}).
      
    \item To support the Open Science movement, we publish the code to our solution and provide easy steps for reproduction/replicability purposes of the experiments (see Sec. \ref{sec:hyperparameters}).
    
\end{enumerate}

\begin{figure*}[!t]
    \centering
    \includegraphics[width=\textwidth]{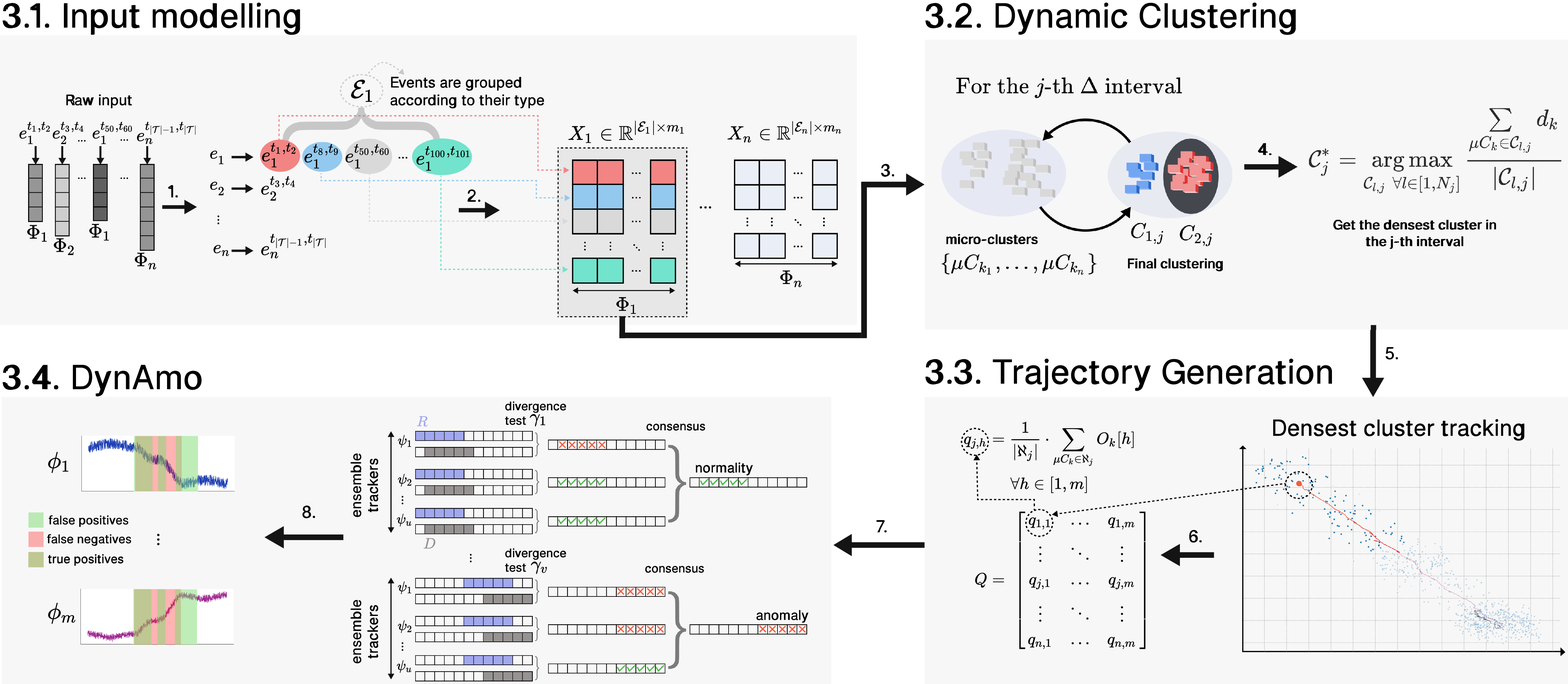}
    \caption{Workflow of the proposed pipeline: i.e. input modelling (described in Sec \ref{sec:input_modelling}), dynamic clustering (Sec. \ref{sec:dynamic_clustering}), trajectory generation (Sec. \ref{sec:trajectory_generation}), and DynAmo (Sec \ref{sec:predictive_strategy}). In \textbf{Step 1}, event vectors, generated in real-time during the signal pre-processing phase, are accumulated according to their type throughout the monitoring period $T$. In \textbf{Step 2}, a sequence $X_i$ is generated for each event type $i$, where each row represents the new event of type $i$ detected along the monitoring period, and each column represents the feature dimension $\phi \in \Phi$. In \textbf{Steps 3} and \textbf{4},  the dynamic clustering component captures an overall time series trend. It produces clusters for each temporal interval $\Delta$ in $T$. The clusters in each interval become the input to the next component - trajectory generation (\textbf{Steps 5} and \textbf{6}) - which extracts features from the densest cluster centroids. Finally, in \textbf{Steps 7} and \textbf{8}, DynAmo predicts the drift areas (green boxes) for each feature of the events in the series, using an ensemble of divergence tests.}
    \label{fig:workflow}
\end{figure*}

\section{Methodology}\label{sec:methodology}

Here, we describe the proposed method for modelling and detecting  anomalous periods in behavioural sequences. To  help the reader understand the model description, Tables \ref{tab:notation_input}, \ref{tab:notation_dyclee}, \ref{tab:notation_trajectory}, and \ref{tab:notation_prediction} summarise the notation used throughout the paper.
\subsection{Application scenario and summary workflow}
\label{sec: scenario and workflow}

We refer to a scenario in which an IoT environment is set to collect signals from a variety of ambient and wearable sensors to monitor specific behaviours such as daily activities (sleeping, eating, personal hygiene), vital signs (pressure, ECG), energy consumption (lighting, heating, use of household appliances),  eating habits, smoking, physical activity, and more. We also assume that one or more sensors are set to monitor a specific behaviour (e.g., for sleep: pressure sensor in the bed, lighting, wearable sleep trackers), generating pre-processed signals transformed into temporal sequences of discrete events (\textit{data points}).  

The problem addressed in this section is detecting gradual changes (drift anomalies) in the characterisation of data points, for example, changes in sleep quality. Note that addressing single behaviours does not mean that they are considered in isolation since, as described hereafter, a behaviour is represented as a complex event that may include contextual features such as, for the case of sleep, the activities performed before and after, or any breaks to go to the toilet.

Figure \ref{fig:workflow} illustrates the steps of the proposed pipeline, summarised in the caption. We begin by presenting a mathematical formalisation of the input model (see Sec. \ref{sec:input_modelling}). As shown in the Figure, the crucial steps of drift detection are based on dynamic clustering \cite{barbosa2016novel} of the data points referring to a given observed behaviour, and next, on capturing the trend of the trajectory by building a  stream of centroids that maximally comprehend the original series.

We briefly describe how dynamic clustering works in Sec. \ref{sec:dynamic_clustering}. Trend capturing is crucial because it eliminates noisy data points (such as outliers) within a specific time window. This strategy allows us to discard data coming from the stream that do not contribute to the overall trend of the original behavioural sequence (see Sec. \ref{sec:trajectory_generation}). Then, we explain our prediction framework (see Sec. \ref{sec:predictive_strategy}).

\subsection{Input modelling}\label{sec:input_modelling}

\begin{table}[!t]
\centering
\caption{Notation used for the input modelling.}
\label{tab:notation_input}
\resizebox{\linewidth}{!}{%
\begin{tabular}{@{}ll@{}}
\toprule
 Notation & Description \\ \toprule
$e_i^{b,f}$ & Event of type $i$ with beginning time $b$ and end time $f$ s.t. $b < f$ \\ \midrule 
 $\Phi_i$ & Feature set of event type $i$, i.e., $\Phi_i = \{\phi_{i,1},\dots,\phi_{i,m_i}\}$ \\ \midrule 
  $m_i$ & The number of features for event type $i$, i.e., $m_i = |\Phi_i|$ \\ \midrule 
  $T$ & End timestamp of the overall monitoring period, starting at $t=0$\\ \midrule 
   $\mathcal{E}_i$ & The events of type $i$ within the monitoring time $T$\\ \midrule
  $X_i$ & \begin{tabular}[c]{@{}l@{}}Sequence of events of type $i$ with their associated feature sets $\Phi_i$\\ordered by the beginning time of each event\end{tabular} \\ \midrule
   $\Delta$ & \begin{tabular}[c]{@{}l@{}}Time unit (e.g., hourly, daily, weekly) expressed in seconds that\\ partitions $X_i$ into consecutive intervals\end{tabular} \\ \midrule
 $b(j)$ & The beginning time of the $j$-th $\Delta$ interval \\ \midrule
 $f(j)$ & The end time of the $j$-th $\Delta$ interval \\ \midrule
  $n$ & \begin{tabular}[c]{@{}l@{}} Number of time intervals from the beginning until the end of\\monitoring according to the $\Delta$ time unit, i.e., $n = \lceil\frac{T}{\Delta}\rceil$ \end{tabular} \\
 \bottomrule
\end{tabular}%
}
\end{table}

Before explaining the proposed method, we provide the reader with a brief formalisation of the behavioural sequences given in input. The input modelling described hereafter applies to the context of multi-sensor monitoring of human behaviour in controlled environments. However, it can be easily extended to input data represented as multivariate temporal trajectories. In this context:
\begin{enumerate}[label=\alph*),topsep=0pt,itemsep=0pt]
    \item One or more sensors may concur in identifying a specific event type $i$ such as sleep, hygiene, or eating;
    \item Each event $e_i$ of type $i$ has a feature set $\Phi_i = \{\phi_{i,1},\phi_{i,2},\dots,\phi_{i,m_i}\}$ such as the duration, the beginning time, or the portion of day (e.g. night, morning) in which the event is captured. Here $m_i$ is the number of features\footnote{Note that $m_i$ depends of the observed event. For example, modelling the activity \textit{sleep} may require  specific fine-grained descriptors in addition to those previously listed, such as the number of interruptions to go to the toilet, the sleep phases (e.g., light vs. deep).}  of the event type $i$;
    \item Events can be non-contiguous since the environment can contain blind spots out of sensor reach, or there might be unobserved/unobservable behaviours.
    \item Events (also of different types) are non-overlapping.
\end{enumerate}
These time series are composed of events of different types with associated beginning and end times. In other words, we consider a discrete time series of events $e_i^{b,f}$ with beginning time $b$ and end time $f$ such that $b < f$, and type $i$. Notice that, in our scenario, two events of any type\footnote{We use $*$ to denote any events in our time series.} $e_*^{b',f'}$ and $e_*^{b'',f''}$ cannot overlap with one another: i.e., $b' \geq f'' \; \lor \; b'' \geq f'$. Additionally, each event $e_i$ has its feature vector $\Phi_i$. Two events of the same type $i$ have the same number of features $m_i$, while two events of different types might be mapped to two different feature vector spaces. 

We group events of type $i$ in a set $\mathcal{E}_i$ and order them according to their beginning time to generate a behavioural representation - see step 1. Thus, we can define $X_i$ as the sequence that contains the feature vectors of events $e_i$ - see step 2 - for the entire monitoring time $T$. To comply with the formalisation in \cite{aragona2021coronna,prenkaj2021hidden,PRENKAJ2023102454}, we reserve two features in $\Phi_i$ to describe the beginning $b$ and end time $f$ of an event $e_i$. Therefore, the temporal dimension of the events is preserved through this input transformation.

Note that our framework is adaptable to multiple types of events and, in particular, it is \textit{completely agnostic with respect to the event type and its number and type of features}, information that is only used to explain a detected drift during subsequent human inspection\footnote{For example, medical personnel are notified that a detected behavioural drift has involved the duration and onset time features of sleep events.}. Therefore, for simplicity and notation readability purposes, we omit the type index $i$ from the notation and assume that all the following formulas and discussions hold for any event type. So, now, $X_i$ becomes $X$, $\Phi_i$ is $\Phi$, $m_i$ is $m$, $e_i^{b,f}$ is $e^{b,f}$, and $\mathcal{E}_i$ is $\mathcal{E}$.

One can view $X \in \mathbb{R}^{|\mathcal{E}| \times m}$ as a real matrix of dimensions $|\mathcal{E}| \times m$. We can divide $X$ into $n$ contiguous intervals with specific duration $\Delta$, i.e., $n = \lceil \frac{T}{\Delta} \rceil$. Suppose each $j$-th $\Delta$ interval s.t. $1 \leq j \leq n$ has a beginning $b(j)$ and end $f(j)$ time. In this way, $X[b(j)$:$f(j)]$ delineates those events that are within the temporal boundaries of the $j$-th $\Delta$ interval. Notice that events spanning over the temporal boundaries $[b(j),f(j)]$ (i.e., their end time exceeds $f(j)$) are split into distinct contiguous events, where the remainder spans over to the next $(j+1)$-th $\Delta$ interval. For example, this could happen when we trace a \textit{sleep} event, which begins at 21:30:00 of day $j$ and ends at 04:00:00 of day $j+1$. Assuming that we are grouping events according to a daily interval (thus, $\Delta=1$ day), we split this event into two: i.e., the first begins at 21:30:00 and ends at 23:59:59 of day $j$, and the second begins at 00:00:00 and finishes at 04:00:00 of day $j+1$. This reasoning can be extended to propagate the remainder of over-spanning events in multiple contiguous $\Delta$ intervals.

\subsection{Capturing trends via dynamic clustering}\label{sec:dynamic_clustering}

\begin{table}[!t]
\centering
\caption{Notation used for capturing trends via dynamic clustering}
\label{tab:notation_dyclee}
\resizebox{\linewidth}{!}{%
\begin{tabular}{@{}ll@{}}
\toprule
 Notation & Description \\ \toprule
 $\mu C_k$ & $\mu$-cluster, i.e., a group of events close in all their feature dimensions $\Phi$ \\ \midrule 
  $p_k$ & Number of events (elements) in $\mu C_k$ \\ \midrule 
  $F_k$ & \begin{tabular}[c]{@{}l@{}}The set of features of each event in $\mu C_k$. It can be represented as a\\ matrix in $\mathbb{R}^{p_k \times m}$ where each row corresponds to the feature set of\\the events in $\mu C_k$\end{tabular} \\ \midrule 
  $\alpha_k$ & The time when $\mu C_k$ is created \\ \midrule 
  $\beta_k$ & The time of last assignment of an event in $\mu C_k$ \\ \midrule 
  $d_k$ & The density of $\mu C_k$ \\ \midrule 
  $O_k$ & The centroid of the $\mu C_k$ cluster \\ \midrule 
  $U_{h}$ &\begin{tabular}[c]{@{}l@{}}The span of the hyperbox of $\mu C_k$  defined as\\$|\max F_k[:,h] - \min F_k[:,h]|$ \end{tabular} \\ \midrule 
  $N_j$ & Number of clusters identified in the $j$-th $\Delta$ interval\\ \midrule 
  $\mathcal{C}_{l,j}$ & \begin{tabular}[c]{@{}l@{}} The $l$-th identified cluster, i.e., a set of connected $\mu$-clusters\\$\{\mu C_{k_1},\dots,\mu C_{k_n}\}$, in the $j$-th $\Delta$ interval s.t. $l \leq N_j$\end{tabular} \\ \bottomrule
\end{tabular}%
}
\end{table}

Unlike other approaches in drift anomaly detection, we use dynamic clustering, a method for tracking evolving environments (see steps 3-4 of Figure \ref{fig:workflow}). We build on top of DyClee \cite{roa2019dyclee} and adapt it to create a trajectory of denser clusters used to classify, without any supervision, a sequence as anomalous or not.

DyClee is a distance and density-based algorithm that handles non-convex and multi-density clustering, working incrementally unsupervised. It uses a two-stage algorithm to produce the final dense set of clusters. First, it collects, processes, and compresses data samples in $\mu$-clusters based on the Manhattan distance. In detail, a $\mu$-cluster $\mu C_k$ is a group of events close in all their feature dimensions $\Phi$. A $\mu$-cluster $\mu C_k$ is a tuple $\mu C_k=(p_k, F_k, \alpha_k, \beta_k, d_k, O_k)$ where:
\begin{itemize}[topsep=0pt,noitemsep]
    \item $p_k \in \mathbb{N}$ is the number of elements in $\mu C_k$.
    \item $F_k \in \mathbb{R}^{p_k \times m}$ the feature matrix for each event in $\mu C_k$.
    \item $\alpha_k$ is the time when $\mu C_k$ was created.
    \item $\beta_k$ is wwhen the last event was assigned to $\mu C_k$.
    \item $d_k$ is the density of $\mu C_k$.
    \item $O_k = \{o_{1},\dots,o_{m}\}$ s.t. $o_{i} = \frac{1}{p_k} \cdot \sum_{j=1}^{p_k} F_k[j,i]$ is the centroid of $\mu C_k$.
\end{itemize}
Notice that each $\mu$-cluster can be seen as an $m$-dimensional vector - hereafter \textit{hyperbox}. This hyperbox is bound in size at each feature dimension according to $U_h = |\max F[$:$,h] - \min F[$:$,h]| \; \forall h \in [1,m]$. Hence, we can express the density of $\mu C_k$ as $d_k = \frac{p_k}{\prod_{h=1}^m U_h}$ where the denominator is the volume of the bounding hyperbox.

Recall that we choose a $\Delta$ time unit to divide $X$ into $n$ contiguous portions. We use $\Delta$ to coordinate the clustering procedure that operates as a wake-up protocol over the incoming events $X$. At each $j$-th $\Delta$ interval, DyClee wakes up with an empty set of $\mu$-clusters. The first event in $X[b(j)$:$f(j)]$ becomes the centre of the first $\mu$-cluster. The other incoming events are grouped according to Def. \ref{def:dyclee_reachable_cluster}. Here, we abuse the original notation of the event to facilitate readability and assume that an event now corresponds to a row in $X$.

\begin{definition}\label{def:dyclee_reachable_cluster}
A $\mu$-cluster $\mu C_k$ is reachable from event $e \in X[b(j)\text{:}f(j)]$ with feature vector $\Phi = \{\phi_1,\dots,\phi_m\}$ if
\begin{align*}
    L_{\infty}(e, \mu C_k) \equiv \max_{h}|\phi_h-O_k[h]|<\frac{U_h}{2}\;\forall h \in [1,m]
\end{align*}
\end{definition}

\noindent and assigned to the closest reachable cluster according to the Manhattan distance
\begin{equation*}
    L_1(e, \mu C_k) = \sum_{h=1}^m | \phi_h - O_k[h]|
\end{equation*}
Notice that with each new event being part of $\mu C_k$ its tuple elements get updated: i.e., $p_k$ is incremented by one, $\beta_k$ reflects a new assignment time, $d_k$ is recalculated, and $O_k$ shifts w.r.t. the new features (row) added to $F_k$.

Having generated the $\mu$-clusters, DyClee enters its second phase (step 4), during which clusters are classified as a $\mu$-cluster w.r.t. its density as dense ($\mathbb{D}\mu$-cluster), semi-dense ($\mathbb{S}\mu$-cluster), or low-dense ($\mathbb{O}\mu$-cluster). Dynamic cluster is important to monitor trend changes in the distribution of $\mu$-cluster, thus, that of the incoming events. To monitor temporal trend shifts of $\mu C_k$, the algorithm employs a forgetting function based on the $j$-th $\Delta$ interval and $\beta_k$. As suggested in \cite{roa2019dyclee}, we use $e^{-0.02(t-\beta_k)}$ where $t$ represents the wake-up time of DyClee in the $j$-th $\Delta$ interval. The forgetting function discards those $\mu$-clusters that are no more significant in determining the trend shifts. DyClee's second phase outputs a set of $\mu$-cluster groups where each $\mu$-cluster in a group is a $\mathbb{D}\mu$-cluster and all its surrounding $\mu$-clusters are either $\mathbb{S}\mu$- or $\mathbb{D}\mu$-clusters. A cluster can be defined according to Def. \ref{def:dyclee_connected_clusters} and \ref{def:dyclee_cluster}. Here, $N_j \in \mathbb{R}_{+}$ denotes the number of clusters identified in the $j$-th $\Delta$ interval.

\begin{definition}\label{def:dyclee_connected_clusters}
\textbf{($\mathbf{\mu}$-cluster connectivity)} $\mu C_{k_1}$ and $\mu C_{k_n}$ are connected if there is a set of $\mu$-clusters $\{\mu C_{k_1},\dots,\mu C_{k_n}\}$ such that the hyperbox of $\mu C_{k_i}$ overlaps with that of $\mu C_{k_{i+1}}$  $\forall i \in [1,n)$ in all but $\theta$ dimensions.
\end{definition}

\begin{definition}\label{def:dyclee_cluster}
\textbf{(dynamic cluster)} A set of $\mu$-clusters $\mathcal{C}_{l,j} = \{\mu C_{k_1},\dots,\mu C_{k_n}\} \text{ s.t. } 1 \leq l \leq N_j$ is the $l$-th identified (dynamic) cluster in the $j$-th $\Delta$ interval if all the $\mu$-clusters are connected with one another.
\end{definition}

\noindent DyClee assumes that low-density $\mathbb{O}\mu$-clusters are outliers in each of the $\Delta$ intervals, an assumption that does not fit the context considered in our study. $\mathbb{O}\mu$-clusters do not necessarily include an anomalous event because their density is lower than the median of the already-formed $\mu$-clusters. Moreover, the construction of $\mathbb{O}\mu$-clusters is capable of capturing only abrupt anomalous events (i.e. point anomalies) and not gradual trend shifts in the trajectory (i.e. drift anomalies). This is because $\mathbb{O}\mu$-clusters can transit into being $\mathbb{S}\mu$-clusters in case their density exceeds the median of the other $\mu$-clusters formed throughout the first step of the algorithm. Therefore, the density of a $\mu$-cluster is a necessary but insufficient criterion in determining anomalous clusters that continuously grow in time due to a drift happening in the original time series.

\subsection{Trajectory generation}\label{sec:trajectory_generation}

\begin{table}[!t]
\centering
\caption{Notation used for the trajectory generation.}
\label{tab:notation_trajectory}
\resizebox{\linewidth}{!}{%
\begin{tabular}{@{}ll@{}}
\toprule
 Notation & Description \\ \toprule
 $\mathcal{C}^*_j$ & The densest cluster in the $j$-th $\Delta$ interval \\ \midrule
 $\mathcal{C}^*[j'$:$j'']$ & \begin{tabular}[c]{@{}l@{}} The set/trace of the densest clusters $\mathcal{C}^*_j$ s.t. $j' \leq j \leq j''$\end{tabular} \\ \midrule 
  $q_{j,h}$ & The centroid of feature $\phi_h \in \Phi$ of $\mathcal{C}^*_j$ \\ \midrule 
  $Q$ & The trend trajectory, i.e., $Q \in \mathbb{R}^{n \times m}$ \\\bottomrule
\end{tabular}%
}
\end{table}

\begin{figure}[!t]
    \centering
    \includegraphics[width=\linewidth]{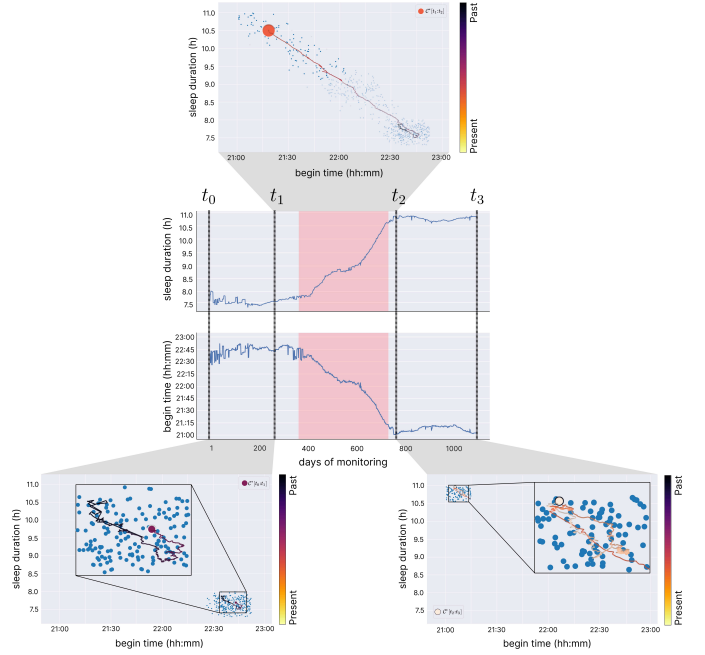}
    \caption{Trajectory generation via the densest cluster in each day. The central part of the image illustrates the sleep pattern of a patient according to duration and begin time. We divide the time series into three time-intervals: i.e. $[t_0,t_1]$ and $[t_2,t_3]$ depicting normality, and $[t_1,t_2]$ depicting a potential shift of distributions. The red area is the ground truth. In $[t_0,t_1]$ (lower-left) the densest clusters $\mathcal{C}^*{[t_0,t_1]}$ are contained in a specific region; in $[t_1,t_2]$ (upper-central) we detect $\mathcal{C}^*{[t_1,t_2]}$ whose trend is shifting towards the upper quadrant of the plot indicating a possible drift; in $[t_2,t_3]$ (lower-right) the trace of $\mathcal{C}^*{[t_2,t_3]}$ is again contained within a specific region, depicting a new stable state.}
    \label{fig:trajectory_generation}
\end{figure}

We add a mechanism to identify trajectory trends from dynamic clusters to account for the drawback mentioned above in signalling a potential drift (see steps 5-6 in Figure \ref{fig:workflow}). We trace the densest cluster for each $j$-th $\Delta$ interval to form a trend trajectory of $X$. We obtain the densest cluster in each $j$-th $\Delta$ interval as follows:
\begin{equation}\label{eq:densest_cluster}
    \mathcal{C}_{j}^* = \underset{\mathcal{C}_{l,j} \;\forall l \in [1,N_j]}{\arg\max}\frac{\sum\limits_{\mu C_k \in \mathcal{C}_{l,j}} d_k}{|\mathcal{C}_{l,j}|}
\end{equation}

\noindent Figure \ref{fig:trajectory_generation} illustrates the trajectory generation process. In this example, we observe a patient's sleep patterns, with $\Delta = 1$ day. The red area in the Figure shows how the patient begins sleeping earlier and longer than usual. For illustration purposes, we divide the monitoring period into three portions:
\begin{itemize}[topsep=0pt,noitemsep]
    \item $[t_0,t_1]$ and $[t_2,t_3]$ depict the normality of the series where the first interval before the drift and the second corresponds to the new behaviour after the drift;
    \item $[t_1,t_2]$ represents a potential shift of feature distributions.
\end{itemize}
For readability purposes, we use $\mathcal{C}^*{[a,b]}$ to indicate the set of densest clusters $\mathcal{C}^*_j$ in the $j$-th $\Delta$ interval s.t. $j \in [a,b]$. In this way, Figure \ref{fig:trajectory_generation} depicts that $\mathcal{C}^*{[t_0,t_1]}$ and $\mathcal{C}^*{[t_2,t_3]}$ remain within a specific region of the plot, while $\mathcal{C}^*{[t_1,t_2]}$ moves diagonally from one extreme to the other. We exploit each densest cluster $\mathcal{C}^*_j$ and trace how its centroid moves during the monitoring time. To this end, we use the centroid of each $\mathcal{C}^*_j$ in the $j$-th $\Delta$ interval, which can be represented as follows:
\begin{equation}
    q_{j,h} = \frac{1}{|\mathcal{C}^*_j|} \cdot \sum_{\mu C_k \in \mathcal{C}^*_j} O_k[h] \;\forall h \in [1,m]
\end{equation}
\noindent where $O_k$ is the centroid of $\mu C_k$. In other words, the centroid of the densest cluster $\mathcal{C}^*_j$ is the mean of the centroids of its $\mu$-clusters on all feature dimensions $h \in [1,m]$. Hence, for each $j \in [1,n]$, we the overall trend trajectory $Q \in \mathbb{R}^{n \times m}$ is
\begin{equation}
Q = 
\begin{bmatrix}
q_{1,1} & \dots & q_{1,m}\\
\vdots & \ddots & \vdots\\
q_{j,1} & \dots & q_{j,m}\\
\vdots & \ddots & \vdots\\
q_{n,1} &\dots & q_{n,m}
\end{bmatrix}
\end{equation}

\subsection{Predictive strategy}\label{sec:predictive_strategy}

\begin{table}[!h]
\centering
\caption{Notation used for the prediction strategy}
\label{tab:notation_prediction}
\resizebox{\linewidth}{!}{%
\begin{tabular}{@{}ll@{}}
\toprule
  Notation & Description \\ \toprule
 $Q[j'$:$j'']$ & \begin{tabular}[c]{@{}l@{}}The trend trajectory containing only the centroids that are within\\the $j'$-th and $j''$-th $\Delta$ intervals s.t. $1 \leq j' \leq j'' \leq n$,\\i.e., $Q[j'$:$j''] = \{q_{j,h} \; | \; q_{j,h} \in Q \; \land \; j' \leq j \leq j'' \; \forall \phi_h \in \Phi\}$\end{tabular} \\ \midrule
  $\delta$ & Time step that the windows move at each iteration \\ \midrule 
  $\ell$ & \begin{tabular}[c]{@{}l@{}}The size used for both reference and detection windows.\\Each window contains $ \frac{\ell}{2} $ hyperboxes\end{tabular} \\ \midrule 
  $\lambda$ & Past hyperboxes considered to reconstruct the evolution trend \\ \midrule 
  $\sigma$ & \begin{tabular}[c]{@{}l@{}} The consensus threshold when aggregating the anomaly detectors\\via trackers\end{tabular} \\ \midrule 
  $\Psi$ & Ensemble of trackers $\Psi = \{\psi_1,\dots,\psi_u\}$ \\ \midrule 
  $\Gamma$ & Set of divergence tests $\Gamma = \{\gamma_1,\dots,\gamma_v\}$ \\ \midrule 
  $g(j',j'')$ & The girth of the hyperbox of the centroids in $Q[j'$:$j'']$ \\ \midrule 
  $\hat{g}(j',j'')$ & The L2-norm of the girth of the hyperbox of the centroids in $Q[j'$:$j'']$ \\ \midrule 
  $G(Q',Q'',\phi_h,f)$ & \begin{tabular}[c]{@{}l@{}}The element-wise difference between the hyperboxes for feature $\phi_i$\\in $Q'$ and $Q''$, aggregated via a generic function $f$,\\i.e., $ f(Q'[\text{:},h]) - f(Q''[\text{:},h])$\end{tabular} \\ \midrule 
  $\hat{G}(Q',Q'',f)$ & \begin{tabular}[c]{@{}l@{}}The L2-norm between the hyperboxes for all features in $Q'$ and $Q''$,\\aggregated via a generic function $f$, i.e., $\big|\big|f(Q')-f(Q'')\big|\big|^2_2$\end{tabular} \\ \midrule 
  $R(\psi)$ & Reference window of hyperboxes traced by tracker $\psi \in \Psi$ \\ \midrule 
  $D(\psi)$ & Detection window of hyperboxes traced by tracker $\psi \in \Psi$ \\ \midrule 
  $\hat{Y}$ & \begin{tabular}[c]{@{}l@{}} Drift output matrix $\hat{Y} \in \{0,1\}^{v\times u}$\\s.t. $\hat{Y}[i,j] = 0 \text{ if  } \gamma_i(R(\psi_j), D(\psi_j))$, $1$ otherwise \end{tabular} \\ \bottomrule
\end{tabular}%
}
\end{table}

In what follows, we illustrate the algorithm to predict behavioural drifts, which is the central contribution of this work. We begin with a high-level summary of the algorithm.
\noindent\textbf{Summary: } In online drift detection scenarios, we use two sliding windows, named \textit{reference} $R$ and \textit{detection} $D$. We signal a drift if their data distributions change according to a divergence test $\gamma$ (e.g., KL divergence) applied on the hyperboxes of $Q$. If $\gamma$ gives a negative outcome, the elements belonging to the detection window become part of the reference window. Both windows move according to each iteration's time step $\delta$. We set the window sizes to $\frac{\ell}{2}$ s.t. $4 \leq \ell \leq \lfloor \frac{n}{2} \rfloor \;\land\;\ell\equiv 0\pmod{2}$. Notice that $\ell$ is a multiple of $\Delta$ intervals. For example, if $\Delta = 1 \text{ day}$, then $\ell$ might be $4$, or $16$ days: i.e., $\ell = 4 \Delta$ and $16\Delta$, respectively.

Unlike other sliding window detectors, DynAmo can reconstruct the evolution trend of the last $\lambda$ hyperboxes. In this way, the distribution change between the reference and detection windows transcends the mere current view of the hyperboxes, including their evolution during the monitoring time. The evolution of the $\lambda$ hyperboxes can be traced in different ways described below. Here, we adopt an ensemble of trackers $\Psi = \{\psi_1,\dots,\psi_u\}$ allowing us to have several detection strategies which lead to a more robust way of detecting anomalies \cite{chawla2001creating}. Similarly, we use a set of divergence tests $\Gamma = \{\gamma_1,\dots,\gamma_v\}$ that generate the divergence prediction for each component $\psi \in \Psi$.

\begin{algorithm}[!ht]
\caption{\texttt{DynAmo}: Ensembles of window-based trackers and drift checkers.}\label{algo:dynamo}
\begin{algorithmic}[1]
\Require$Q \in \mathbb{R}^{n\times m}$, $\delta > 0$, $0 \leq \lambda < n$, $4 \leq \ell \leq  \lfloor \frac{n}{2} \rfloor \;\land\;\ell\equiv 0\pmod{2}$, $0 < \sigma < 1$, $\Psi = \{\psi_1,\dots,\psi_u\}$, $\Gamma = \{\gamma_1,\dots,\gamma_v\}$
\State $\hat{y} \gets [0,\dots,0] \text{ s.t. } |\hat{y}|=n$
\State $t \gets 1$
\State $R, D \gets \{\psi:\emptyset\; \forall \psi \in \Psi\}, \{\psi:\emptyset\; \forall \psi \in \Psi\}$
\While{$t \leq n$\tikzmark{G}}
    \If{$n-t+1 < \ell$}
        \State $\ell \gets \lfloor\frac{n-t+1}{2}\rfloor$
    \EndIf\tikzmark[xshift=5.7cm]{H}
    \State\tikzmark{A} $W_{prev} \gets \emptyset$
    \For{$j=t \text{\textbf{ to }} t +  \frac{\ell}{2}  - 1$}
        \State $W_{curr} = Q[\max\{1,j-\lambda\}$:$j]$
        \For{$\psi \in \Psi$}
            \State $R[\psi].\text{\texttt{add}}(\psi.\text{\texttt{track}}(W_{prev}, W_{curr}))$
        \EndFor
        \State $W_{prev} \gets W_{curr}$
    \EndFor\tikzmark[xshift=5.5cm]{B}
    \State\tikzmark{C}$W_{prev} \gets \emptyset$
    \For{$j = t +  \frac{\ell}{2}  \text{\textbf{ to }} t + \ell - 1$}
        \State $W_{curr} \gets Q[\max\{1,j-\lambda\}$:$j]$
        \For{$\psi \in \Psi$}
            \State $D[\psi].\text{\texttt{add}}(\psi.\text{\texttt{track}}(W_{prev}, W_{curr}))$
        \EndFor
        \State $W_{prev} \gets W_{curr}$
    \EndFor\tikzmark[xshift=5.5cm]{D}
    \State$\hat{Y} \gets \text{\texttt{detect}}(R,D,\Gamma)$\tikzmark{E}
    \If{\texttt{consensus}$(\hat{Y}, \sigma) = 1$}
        \State $\hat{y}[t+\frac{\ell}{2}$:$t+\ell-1] = 1$
        \State $t \gets t + \frac{\ell}{2}$
    \Else
        \State $t \gets t + \delta$
    \EndIf
    \State\tikzmark[xshift=6.6cm]{F}$R, D \gets \{\},\{\}$
\EndWhile
\State \Return $\hat{y}$
\end{algorithmic}
\end{algorithm}

\begin{tikzpicture}[remember picture,overlay]
\draw[black,thick,decorate,decoration={brace,amplitude=5pt,mirror}](B.north)--node[right=.1em]{\rotatebox{270}{\textit{Fill reference wnd.}}}(B|-A);
\end{tikzpicture}

\begin{tikzpicture}[remember picture,overlay]
\draw[black,thick,decorate,decoration={brace,amplitude=5pt,mirror}](D.north)--node[right=.3em]{\rotatebox{270}{\textit{Fill detection wnd.}}}(D|-C);
\end{tikzpicture}

\begin{tikzpicture}[remember picture,overlay]
\draw[black,thick,decorate,decoration={brace,amplitude=5pt,mirror}](F.north)--node[right=.1em]{\rotatebox{270}{\textit{Is there a drift?}}}(F|-E);
\end{tikzpicture}

\begin{tikzpicture}[remember picture,overlay]
\draw[black,thick,decorate,decoration={brace,amplitude=5pt,mirror}](H.north)--node[right=.1em]{\rotatebox{270}{\textit{Is $\ell$ too big?}}}(H|-G);
\end{tikzpicture}

\begin{algorithm}[!t]
\caption{\texttt{detect}: Subroutine to detect the drift in the two windows.}\label{algo:detector}
\begin{algorithmic}[1]
\Require $R, D, \Gamma = \{\gamma_1,\dots,\gamma_v\}$
\State $\hat{Y} \gets 0^{v \times u}$
\For{$t, \gamma \in \text{\texttt{enumerate}}(\Gamma)$}
    \For{$j, \psi \in \text{\texttt{enumerate}}(R.\text{\texttt{keys}}())$}
        \State $\hat{Y}[t,j] \gets \neg\;\gamma(R[\psi],D[\psi])$
    \EndFor
\EndFor
\State \Return $\hat{Y}$
\end{algorithmic}
\end{algorithm}

To put this in perspective, we have $u \times v$ predictions at each iteration. By relying on the look-back hyperparameter $\lambda$, we track how the hyperbox of the densest cluster evolves in time. This allows DynAmo to track the impact of the "movement" of the centroid corresponding to the densest cluster (see Fig. \ref{fig:trajectory_generation}) in time. Hence, DynAmo can interpret slight evolution shifts in $Q$. Additionally, DynAmo can identify drift periods as they are happening because, in case of drifts, the old detection window becomes the new reference window, thus updating the current view of "normality". As depicted in Fig. \ref{fig:trajectory_generation}, the drift interval shows a continuous change in the data distribution. DynAmo's windows go through the drift interval in these scenarios, constantly firing drift signals. Lastly, we aggregate the prediction of all $\gamma \in \Gamma$ for each tracker $\psi \in \Psi$. Notice that DynAmo is a flexible framework that accepts any set of trackers, divergence tests, and aggregation functions (e.g., majority voting) that might use a threshold $\sigma$ for binarising the predictions.

\noindent
\textbf{Algorithm description:} 
To help the reader understand the algorithm \ref{algo:dynamo}, we first use $Q[j'$:$j'']$ to denote the set of centroids in $Q$ that are within the $j'$-th and $j''$-th $\Delta$ intervals. We invite the reader to imagine the population of the reference and detection window as an incremental  procedure to track how the centroids in $Q$ evolve with each passing iteration (see the upper subplot in Fig. \ref{fig:trajectory_generation}). Specifically, we want to track how the $m$-dimensional feature vector (hyperbox) of each centroid changes with each passing $j$-th $\Delta$ interval. Each window, $R$ and $D$, will contain exactly $\frac{\ell}{2} \times u$ tracked hyperboxes in each iteration. The procedure is in four steps, as indicated in Algorithm \ref{algo:dynamo} and detailed hereafter.

\noindent\textit{1. Fill the reference window}: First, we fill the reference window (lines 6-13). Here, we use an alternating and incremental strategy to keep track of the evolution of the hyperboxes that contain the centroids in a specific time interval. Therefore, we can rely on $Q[j'$:$j'']$ to slide through the time series. In particular, for the reference window, we need to see how adding a new centroid (line 10) changes the bounding hyperbox (line 12). Since we can look at $\lambda$ steps in the past, we clamp the left extreme such that it does not go past the first $\Delta$ interval. Assuming that we do not fall off bounds, in each iteration, $W_{curr}$ contains the centroids from $j-\lambda$ to $j$, whereas $W_{prev}$ contains those from $j-\lambda-1$ to $j-1$. In this way, tracking how $W_{curr}$ changes w.h.t. $W_{prev}$ means measuring how much the centroids' distribution changes if we "forget" the first one in $W_{prev}$ and consider the one corresponding to the $j$-th $\Delta$ (current) interval. Hence, with each new centroid being added to $W_{curr}$, we track its impact on the evolution of the hyperbox in $W_{prev}$ (line 12). \textit{We remark that this is one central idea of DynAmo, which makes it robust to oscillatory signals and sensible to slight changes, as later shown in Sec. \ref{sec:experiments}}. Next, we update the reference window $R$ represented as a dictionary of key-value pairs for each tracker $\psi \in \Psi$. Notice that $R$ contains the evolution for each batch of $\frac{\ell}{2}$ incoming centroids. Because we use an ensemble, we track the evolution from $W_{prev}$ to $W_{curr}$ for each $\psi \in \Psi$. We assume that each $\psi$ has a method \texttt{track} that measures how $W_{curr}$ changes w.r.t. $W_{prev}$.

\noindent\textit{2. Fill the detection window}: Next, we fill the detection window (lines 16-23). The same reasoning with the reference window can be applied in this case. Notice that $D$ operates on the next $\frac{\ell}{2}$ centroids (line 17). Differently from the reference window, when filling the detection window with the tracking result on $W_{prev}$ and $W_{curr}$, we allow it to look $\lambda$ steps in the past without worrying that this looking back might lead beyond the limits of the window (line 18). In other words, we allow looking back into what might have happened in the reference window instead of cutting it to fit the temporal interval of the detection window from the left side (i.e., line 18 would be $W_{curr} \leftarrow Q[\max\{t+\frac{\ell}{2},j-\lambda\}$:$j]$). This is a desired property since the detection window might contain a portion of evolutive history outside its temporal bounds to track how the current trend of the centroid hyperboxes behaves w.h.t. the past (see upper subplot in Fig. \ref{fig:trajectory_generation}). Similarly to $R$, $D$ exploits the same data structure containing key-value pairs for each tracker $\psi \in \Psi$.

\noindent\textit{3. Is there a drift?} Lastly, when we have populated both $R$ and $D$ with the $\ell$, we need to detect whether $D$ contains a drift. We rely on Algorithm 1 to output a matrix $\hat{Y} \in \{0,1\}^{v\times u}$ that contains the result of each divergence test $\gamma \in \Gamma$ on the reference and detection windows. Notice that we use the utility function \texttt{enumerate} to loop through an array by accessing simultaneously the current element and its index (e.g., lines 2-3). Additionally, we assume that $R$ and $D$, dictionaries, have a function called \texttt{keys} which returns their key sets, i.e., $\Gamma$. Hence, for each divergence test $\gamma$, we check if it gives a negative result on the tracked hyperboxes $R[\psi]$ and $D[\psi] \; \forall \psi \in \Psi$. If the divergence test gives a negative result, then there is a drift, and we set the $i,j$ cell of $\hat{Y}$ to $1$; otherwise, to $0$ (line 4). This subroutine returns the drift-decision matrix $\hat{Y}$ to line 24 in Algorithm \ref{algo:dynamo}. Subsequently, since we have multiple decisions - i.e., $v \times u$ - we aggregated them according to a function \texttt{consensus}$: \{0,1\}^{v \times u} \times \mathbb{R} \rightarrow \{0,1\}$ (line 25) which might use a binarisation threshold $\sigma$. Suppose the aggregation of decisions suggests that there is drift. In that case, we label the current window anomalous (line 26) and replace the reference window with the detection\footnote{From the pseudocode, one can notice that we do not explicitly replace $R$ with $D$. Instead, we move $i$ to the index corresponding to where the current detection window starts, i.e., $i + \frac{\ell}{2}$. Therefore, in the next iteration, $R$ will be filled with the hyperboxes of $D$ in the previous iteration.} (line 27). If no drift is signalled (line 28), we move both windows forward by $\delta$ time steps. Notice that, in false positive cases - i.e., identifying an inexistent drift - DynAmo wrongly replaces $R$ with $D$. However, in the next iteration, $R$ will contain elements that, in reality, are normal, thus not hindering the correct functioning of the algorithm. \textit{The ensemble-based prediction strategy outlined in this paragraph is another relevant aspect of DynAmo that contributes to its stability across different datasets and anomaly types (see Sec. \ref{sec:experiments})}.

\noindent\textit{4. Is $\ell$ too big?} At each iteration, we reset both $R$ and $D$. To ensure the algorithm does not exceed the $n$-th $\Delta$ interval, we check for enough space to fill the last two windows (line 5) and redefine $\ell$ as half of what remains. The algorithm might suffer from one-off - i.e., the last centroid remaining - if $n\equiv 1\pmod{2}$. However, we ignore this edge case here for clarity.

\noindent\textbf{Dynamo's components:} Notice that DynAmo can take, in a plug-and-play fashion, any tracker, divergence test, and consensus function without modifying its backbone. For reproducibility, we describe what we used in this paper.

\noindent\textit{Trackers}: We use four hyperbox trackers, i.e., $\Psi = \{\psi_1,\psi_2,\psi_3,\psi_4\}$. We rely on the notation in Table \ref{tab:notation_prediction} to briefly describe them here.
\begin{itemize}
    \item $\psi_1$ tracks the volume of the hyperbox, $g(j',j'')$, within the bounds $[j',j'']$ \[g(j',j'') = \prod_{h = 1}^{m} \big(\overbrace{\max Q[j'\text{:}j''][\text{:},h]}^\text{$\max$ value for $\phi_h \in \Phi$} - \overbrace{\min Q[j'\text{:}j''][\text{:},h]}^\text{$\min$ value for $\phi_h \in \Phi$}\big)\] Here, we first select the centroids within $[j',j'']$. Then, we calculate the span (i.e., difference between maximum and minimum) of each feature $\phi_h \in [1,m]$.
    \item $\psi_2$ tracks the L2-norm of the hyperbox, $\hat{g}(j',j'')$, within the bounds $[j',j'']$ $$\hat{g}(j',j'') = \sqrt{\sum_{h=1}^m \big(\max Z - \min Z\big)^2}$$ where $Z$ here is a short-hand for $Q[j'\text{:}j''][\text{:},h]$.
    \item $\psi_3$ tracks the difference between two hyperboxes in terms of minimum (maximum) spans. First, we use two temporal bounds $[x,y]$ and $[w,z]$ to get 
    $Q' = Q[x\text{:}y]$ and $Q'' = Q[w\text{:}z]$. In other words, we calculate the difference between the maximum value of $\phi_h \in \Phi$ in $Q'$ and that for the same feature in $Q''$. Hence, we denote with $G(Q',Q'',\phi_H, f) = f(Q'[\text{:},h]) - f(Q''[\text{:},h])$, where $f$ is an aggregation function. The difference between the two hyperboxes on all elements is the matrix
    \begin{equation*}
        \begin{bmatrix}
        G(Q',Q'',\phi_1,\max)  & G(Q',Q'',\phi_1,\min)\\
        \vdots  & \vdots \\
        G(Q',Q'',\phi_m,\max) & G(Q',Q'',\phi_m,\min)
        \end{bmatrix}
    \end{equation*}
    Interestingly, the sign in these vectors illustrates shrinkages/expansions happening for each feature $\phi_h$ through time; meanwhile, its magnitude is represented by the difference value.
    \item $\psi_4$ tracks the L2-norm of the difference between two hyperboxes in terms of minimum (maximum) spans. Similarly to $\gamma_3$, we use $Q'$ and $Q''$. Here, we calculate
    \begin{equation*}
        \begin{gathered}
            \hat{G}(Q',Q'',\max) = \sqrt{\sum_{h=1}^m (\max Q'[\text{:},h] - \max Q''[\text{:},h])^2}\\
            \hat{G}(Q',Q'',\min) = \sqrt{\sum_{h=1}^m (\min Q'[\text{:},h] - \min Q''[\text{:},h])^2}
        \end{gathered}
    \end{equation*}
    and use these two values for tracking purposes.
\end{itemize}

\noindent\textit{Divergence tests}: We use two different drift detection criteria, i.e., $\Gamma = \{\gamma_1,\gamma_2\}$. Notice that we can rely on well-established divergence tests, such as KL divergence or the Kolmogorov-Smirnov test. Instead, we devise ad-hoc detection criteria. Recall that we use the divergence test once the reference and detection windows (line 24 in Algorithm \ref{algo:dynamo}). Therefore, for each tracker $\psi \in \Psi$, the following divergence test get in input the values of $R[\psi]$ and $D[\psi]$:
$$\gamma_1(\mathbf{x},\mathbf{y}) = \mathbbm{1}\bigg[\sum_{j=2}^{\frac{\ell}{2}} \big| \mathbf{x}[j] - \mathbf{x}[j-1] \big| \geq \sum_{j=2}^{\frac{\ell}{2}}\big| \mathbf{y}[j] - \mathbf{y}[j-1] \big|\bigg]$$
$$\gamma_2(\mathbf{x},\mathbf{y}) = \mathbbm{1}\bigg[\mu(\mathbf{x}) - \sigma(\mathbf{x}) < \mu(\mathbf{y}) < \mu(\mathbf{x}) + \sigma(\mathbf{x}) \bigg]$$
where $\mu(\cdot)$ and $\sigma(\cdot)$ are the mean and standard deviation of the input vector, respectively. For trackers $\psi_3$ and $\psi_4$, which do not output a single value, $\gamma_1$ and $\gamma_2$ check whether their conditions are met in any dimension. Notice that DynAmo's backbone allows prospective users to specify established divergence tests to assess whether the distribution of $R$ changes w.h.t. $D$. Additionally, by employing, for instance, the Kolmogorov-Smirnov test, one can measure the KS statistic, which entails the magnitude of the difference. In this way, one could incorporate several severity levels of drift detection, e.g., small, noticeable, and complete behavioural shifts. In this way, system users can fine-tune DynAmo according to the specific needs of the monitored patient's and caregivers' decisions. This investigation remains as future work.

\noindent\textit{Consensus function}:  We use the average voting aggregation, i.e. if the average of $\hat{Y}$ exceeds $\sigma$, then a drift is signalled; otherwise, it is considered normal behaviour.

\noindent
\textbf{Complexity:} Algorithm \ref{algo:dynamo} converges since it is bound to the last $n$-th $\Delta$ interval. DynAmo does a single pass over the time series. It ensures correct tracking of the evolution of the hyperboxes in the two windows. Again, our strategy remains unsupervised, requiring only $\ell$ data points to build the two windows for detection purposes.
\noindent Here, we assume that the cost for calculating maxima, minima, norms, summations, products and aggregations via consensus functions is $O(1)$. Hence, the time complexity of Algorithm \ref{algo:dynamo} is $O(n\times (\ell u + u v))$. Knowing that $\ell$ can be at most $\frac{n}{2}$, we have that  $O(n\times (\ell  u + u v)) = O(n\times(n  u + uv))$. Generally, $u$ and $v$ can be disregarded since $u << n$ and $v << n$. Therefore, the overall cost of the algorithm is $O(n^2)$. Specifically, the whole algorithm is encapsulated in a loop (line 4), which costs $O(n)$. Populating the two windows (lines 9-15 and 17-23) costs $O(\ell u)$. Lastly, the \texttt{detect} subroutine costs $O(uv)$. The space complexity of Algorithm \ref{algo:dynamo} is $O(n + \ell u + \ell + u  v) = O(n)$. In detail, $O(n)$ accounts for maintaining the vector $\hat{y}$; $O(\ell u)$ for the reference $R$ and detection $D$; $O(\ell)$ for both $W_{curr}$ and $W_{prev}$; and $O(uv)$ for the output matrix $\hat{Y}$ of the \texttt{detect} subroutine.

\section{Datasets}\label{sec:dataset}

\begin{table*}[!t]
\centering
\caption{The dataset characteristics for each monitoring scenario in E-Linus (obtained from real data with realistic perturbations) and the other synthetic datasets. For simplicity purposes, we analyse only the activity of sleep.}
\label{tab:dataset_characteristics}
\setlength\extrarowheight{2pt}
\resizebox{\textwidth}{!}{%
\begin{tabular}{@{}lllccccc|cc@{}}
\toprule
\multicolumn{3}{l|}{Datasets} &
  \begin{tabular}[c]{@{}c@{}}Monitoring\\days\end{tabular} &
  \begin{tabular}[c]{@{}c@{}}\% of drift\\in series\end{tabular} &
  \begin{tabular}[c]{@{}c@{}}Avg. daily duration\\of sleep (h)\end{tabular} &
  \begin{tabular}[c]{@{}c@{}}Avg. daily duration of \\sleep interruptions (mins)\end{tabular} &
  \begin{tabular}[c]{@{}c@{}}When does the patient go\\to sleep on avg.? (hh:mm:ss)\end{tabular} &
  \begin{tabular}[c]{@{}c@{}}Avg. duration of sleep\\during drift (h)\end{tabular} &
  \begin{tabular}[c]{@{}c@{}}Avg. interruptions of sleep\\during drift (mins)\end{tabular} \\ \midrule
\multirow{4}{*}{Real} &
  \multirow{2}{*}{ELP1} &
  \multicolumn{1}{c|}{D} &
  1,460 &
  40.00\% &
  8.96 $\pm$ 1.24 &
  9.22 $\pm$ 3.76 &
  \multicolumn{1}{c|}{22:04:36 $\pm$ 00:37:54} &
  9.22 $\pm$ 0.78 &
  9.17 $\pm$ 3.69 \\
 &
   &
  \multicolumn{1}{c|}{I} &
  1,460 &
  40.00\% &
  7.54 $\pm$ 0.22 &
  12.76 $\pm$ 8.05 &
  \multicolumn{1}{c|}{22:45:24 $\pm$ 00:09:01} &
  7.52 $\pm$ 0.25 &
  15.19 $\pm$ 6.61 \\\cline{2-10}
 &
  \multirow{2}{*}{ELP2} &
  \multicolumn{1}{c|}{D} &
  1,460 &
  40.00\% &
  8.46 $\pm$ 1.51 &
  22.90 $\pm$ 11.82 &
  \multicolumn{1}{c|}{21:35:57 $\pm$ 00:57:31} &
  8.73 $\pm$ 1.18 &
  23.06 $\pm$ 12.07 \\
 &
   &
  \multicolumn{1}{c|}{I} &
  1,460 &
  40.00\% &
  6.89 $\pm$ 0.93 &
  35.84 $\pm$ 18.93 &
  \multicolumn{1}{c|}{22:16:26 $\pm$ 00:43:56} &
  6.86 $\pm$ 0.95 &
  39.61 $\pm$ 16.82 \\ \midrule
\multirow{3}{*}{Synthetic} &
  \multicolumn{2}{l|}{PH} &
  170 &
  47.06\% &
  7.51 $\pm$ 1.37 &
  5.84 $\pm$ 20.86 &
  \multicolumn{1}{c|}{22:19:14 $\pm$ 00:50:32} &
  7.08 $\pm$ 1.72 &
  10.86 $\pm$ 28.75 \\
 &
  \multicolumn{2}{l|}{AS} &
  171 &
  47.37\% &
  7.31 $\pm$ 1.54 &
  4.64 $\pm$ 20.94 &
  \multicolumn{1}{c|}{22:23:39 $\pm$ 00:39:54} &
  6.58 $\pm$ 1.84 &
  7.71 $\pm$ 28.86 \\
 &
  \multicolumn{2}{l|}{VK} &
  151 &
  39.33\% &
  7.97 $\pm$ 3.37 &
  9.60 $ \pm$ 68.16 &
  22:38:19 $\pm$ 00:49:29 &
  9.11 $\pm$ 5.27 &
  25.31 $\pm$ 108.77 \\ \bottomrule
\end{tabular}%
}
\end{table*}

To the best of our knowledge, there are no works in the literature that evaluate datasets containing behavioural trajectories annotated with drifts during the time of monitoring. Rather, the literature concentrates on activity recognition in smart home environments \cite{alemdar2013aras,cook2010learning,tapia2004activity,van2010activity}. However, we rely on the reproduction of ARAS \cite{alemdar2013aras}, VanKastereen \cite{van2010activity}, and PolimiHouse as proposed in \cite{masciadri2018disseminating} to evaluate DynAmo. These datasets have a synthetically generated drift period that we attach to the end of the normal period. Each simulation day contains activities of daily living and each sensor's scheduling timetable (wake-up call). Each dataset comprises 90 days of a virtual inhabitant's life and has drift periods compatible with dementia symptoms.

Additionally, we use the E-Linus (EL) dataset consisting of the daily routines of two patients with symptomatic senile social isolation disorders. We created this dataset by collecting activity data within an ambient assisted living environment for older people during an industry-driven project, as detailed in \cite{PRENKAJ2023102454}. Although two patients may seem small, we consider them as single datasets that present different anomaly types (duration, sequence, start time, daily frequency) on six different activities (sleep, hygiene). Moreover, the challenge here is to learn a model of normality and abnormality tailored to each patient's peculiarities since the conditions of the two selected patients and their habitual activities were very different, for example, one with regular and the other with deregulated sleep patterns.

Given the relatively short monitoring period, we artificially extend these sequences over longer periods, based on small realistic perturbations of the observed routines relying on the tool proposed in \cite{podo2022anomalybyclick}. Furthermore, we injected various types of drifts by perturbating the features according to well-defined rules specified with the help of geriatricians participating in the E-Linus project. For each patient (ELP1 and ELP2), we generated two datasets: D, with perturbations on the sleep duration, and I, with perturbations on the number of sleep interruptions. We describe the feature processing in Sec. \ref{sec:feature_extraction}. Notice that, in this paper, we only consider \textit{sleep} events for all datasets.

Table \ref{tab:dataset_characteristics} illustrates the characteristics of the datasets for \textit{Duration (D)} and \textit{Interruptions (I)}. Notice how the first patient (P1) in EL has more regular sleep patterns than the second patient (P2) - see 3$^\text{rd}$ and 4$^\text{th}$ column of the table. Additionally, we report the average duration of sleep and the interruption duration when the drift occurs (see the last two columns). 

We invite the reader to notice that the synthetic datasets clearly define a much easier scenario to detect abnormal periods of activity. In particular, for PolimiHouse (PH) and ARAS (AS), the sleeping duration decrements noticeably, and the interruptions take approximately twice as much as in the normal period. In VanKastareen (VK), the duration of interruptions and sleep increase, respectively, of $\sim\hspace{-.4em}164\%$ and 14\%. Therefore, for the synthetic datasets, we expect that approaches based on a fixed reference window will have an advantage over others because the drift period happens near the end of the monitoring time. Contrarily, for EL, we expect the fixed-reference window approach to underperform w.r.t. sliding window approaches since the distribution of the sleeping patterns continues to change inside the drift period. 

Additional analyses and discussions on the diversity between the two types of datasets are in Sec. \ref{sec:datasets_characteristics}.

\section{Experiments}\label{sec:experiments}
Here, we describe the experiments performed on all datasets and compare DynAmo with several SOTA systems. Sec. \ref{sec:hyperparameters} enlists the compared methods and explains the experimental and hyperparameter settings to endorse reproducibility for future research. Sec. \ref{sec:discussion} describes the performances of the compared methods and provides detailed insights and limitations of each of them.

\subsection{Compared methods, experimental setup, metrics, and hyperparameters}\label{sec:hyperparameters}
\textbf{Compared methods}: We compare with baseline strategies such as Keep It Simple (KIS)\footnote{Naive baseline that labels every day randomly.}, BinSeg \cite{fryzlewicz2014wild}, BottomUp \cite{keogh2001online}, PELT \cite{killick2012optimal,Wambui2015ThePO}, Window \cite{truong2020selective}, IKSS-bdd \cite{dos2016fast}, IKSSW\footnote{A sliding window approach that extends IKS-bdd.}, and KernelCPD \cite{arlot2019kernel}. We also compare with the following state-of-the-art methods\footnote{We searched for the implementation of all the papers in Table \ref{tab:drift_detector_types}, however, \cite{haque2016sand,costa2018drift,pinage2020drift} are not replicable/reproducible. Whereas, \cite{kim2017efficient,koh2016cd,lughofer2016recognizing,bashir2017framework,sethi2018handling,li2019faad,de2019learning} do not have a publicly available code repository.}: KLCPD \cite{DBLP:conf/iclr/ChangLYP19}, MD3 \cite{sethi2015don}, MD3-EGM \cite{sethi2017reliable}, STUDD \cite{cerqueira2022studd}, D3 \cite{gozuaccik2019unsupervised}, NN-DVI \cite{liu2018accumulating}, ERICS \cite{haug2021learning}, and CDLEEDS \cite{DBLP:conf/cikm/HaugBZK22}. We refer the reader to Sec. \ref{sec:related} and Table \ref{tab:drift_detector_types} for a summary description of compared methods and to Sec. \ref{sec:soa_comparison_with_other_metrics} for more details.

\noindent\textbf{Hyperparameters and reproducibility of DynAmo:} We do not divide the input trajectory into sets for training, validation, and testing to maintain a fully unsupervised drift detection approach. Instead, we label the windows in an online fashion. We use a $\Delta = 1 \text{ day}$ to generate $Q$. We performed a Bayesian optimisation with an average F1 score as the target function - see Sec. \ref{sec:dynamo_hyperparamters} for more details - for 100 trials and achieved the best performances for:
\begin{itemize}
    \item \textit{EL} by setting $\lambda = 25$, $\delta = 10$, $\ell = 30$, and $\sigma = 0.2666$,
    \item \textit{Synthetic datasets (PH, AS, VK)} by setting $\lambda = 4$, $\delta = 4$, $\ell = 16$, and $\sigma = 0.3422$.
\end{itemize}
Notice that optimising for maximising the F1 score does not compromise DynAmo's unsupervised nature since this helps optimise the hyperparameters, not the model parameters. The optimisation strategy could be seen as a grid search on all possible combinations of DynAmo's hyperparameters, where we report the best variation based on the average F1 score. There are no learned/updated parameters in DynAmo in this process since the clustering procedure is frozen and only the two-window (i.e., reference and detection window) procedure is executed.

\noindent\textbf{Fair comparison policy}: To make sure that all of the compared methods are in the same realistic scenario - i.e., a situation in which it is not guaranteed that the reference (training) window represents normality, nor it is known whether any drifts would start to appear during this window - we need to uniform the amount of "history" that each method can consider before making predictions. This phenomenon is even more pronounced with (semi)supervised strategies that need a minimum amount of data points reserved for training purposes. It is natural that we reserve only $\frac{\ell}{2}$ data points for the compared methods for "training" purposes such that they have the same view of the distributional changes as DynAmo has in each iteration. Notice that in STUDD, MD3, MD3-EGM, and D3, this is inherently inhibited due to the fact that the underlying binary classifiers require to have at least one observation belonging to the anomalous class. Therefore, these strategies have an advantage and cannot be fully aligned to fairly compare against the others. Finally, we also optimised the hyperparameters of all SoA methods, similar to what we did for DynAmo. We refer the reader to Sec. \ref{sec:compared_methods} for more details.

\noindent \textbf{Evaluation metrics:} We use average F1 scores over 30 runs to evaluate all the methods. For completeness, in Sec. \ref{sec:soa_comparison_with_other_metrics}, we also report the false positive rate (FPR) and false negative rate (FNR) for all methods in Table \ref{tab:best_performances}. Concerning these latter performance indicators, we argue that in real-world clinical scenarios,  false negative errors should be worse since a "real anomalous" behavioural shift has happened, and DynAmo missed it, leading to potential risks for the monitored patients. Balancing the cost of false positives for the hospital and the risk to patient's health arising from false negatives depends on local policies. However, according to ethical principles, we should assign more weight to human health.

\noindent\textbf{The code to replicate/reproduce our solution and experiments is available online}\footnote{\url{https://github.com/bardhprenkaj/dynamo}}. 

\begin{table}[!t]
\centering
\caption{Average F1 scores (over 30 runs) of SoA methods against DynAmo. Notice that $\times$ represents no convergence. A value is in bold if it is the highest value on average; it is underlined when it is not significantly diverse from the best-performing model according to a one-way ANOVA test with a post-hoc Tukey HSD with a p-value of $0.05$. A value is italic if it is the second-best value on average. S denotes supervised, SS semi-supervised, and U unsupervised learning.}
\label{tab:best_performances}
\resizebox{\linewidth}{!}{%
\begin{tabular}{@{}lclccccccc@{}}
\toprule
\multicolumn{3}{l}{}              & \multicolumn{3}{c}{Synthetic datasets}           & \multicolumn{4}{c}{Realistic datasets}                        \\ \cmidrule(lr){3-9}
\multicolumn{3}{l}{\multirow{-2}{*}{}} &
  PH &
  AS &
  VK &
  ELP1-D &
  ELP1-I &
  ELP2-D &
  ELP2-I \\ \midrule
\multicolumn{1}{l|}{} & U &
  KIS &
  0.4836 & 0.4832 &
  0.4321 &
  0.4426 &
  0.4446 &
  0.4426 &
  0.4421  \\
\multicolumn{1}{l|}{} & U & Pelt \cite{killick2012optimal,Wambui2015ThePO}      & 0.1489 & 0.1505                  & 0.1270     & 0.1979 & 0.1956       & 0.2093 & 0.2093 \\
\multicolumn{1}{l|}{} & U & BinSeg \cite{bai1997estimating,fryzlewicz2014wild}   &  0.1573 & 0.1348 & 0.1422     & 0.2148 & 0.2057       & 0.2119 & 0.2119   \\
\multicolumn{1}{l|}{} & U & Window \cite{truong2020selective}   & 0.0952 & 0.0244 & 0.0656     & 0.0134 & 0.0168       & 0.0231 & 0.0166      \\
\multicolumn{1}{l|}{} & U & BottomUp \cite{keogh2001online}  &  0.1915 & 0.1720 & 0.1875     & 0.1961 & 0.1938       & 0.2093 & 0.2093        \\
\multicolumn{1}{l|}{} & U& IKSSW   & 0.2247 & 0.2222 & 0.0000     & 0.0487 & 0.0000       & 0.0714 & 0.0000      \\
\multicolumn{1}{l|}{\multirow{-6}{*}{\rotatebox{90}{Baselines}}} & U &
  IKSS-bdd \cite{dos2016fast} &
  0.2247 &
  0.6667 &
  0.0000 &
  0.1595 &
  0.0000 &
  0.2176&
  0.0000 \\
  \multicolumn{1}{l|}{} & U & KernelCPD \cite{arlot2019kernel} & 0.5564 & 0.5246 & 0.5316     & 0.4455 & 0.4455       & 0.4484 & 0.4484      \\
  \midrule
\multicolumn{1}{l|}{} & S & MD3 \cite{sethi2015don}  & 0.7018 & \textbf{0.9051} & 0.0000     & 0.0000 & 0.6273 & 0.6282 & 0.6525 \\
\multicolumn{1}{l|}{} & SS & MD3-EGM \cite{sethi2017reliable}  & 0.0282                     & 0.0278 & 0.0392     & 0.0036 & 0.0036       & 0.0036 & 0.0036   \\
\multicolumn{1}{l|}{} & SS & STUDD \cite{cerqueira2022studd}    & 0.0909                     & 0.0000 & 0.0000     & 0.6708 & 0.0315 & 0.3068 & 0.1250       \\
\multicolumn{1}{l|}{} & U & KLCPD \cite{DBLP:conf/iclr/ChangLYP19}    & 0.1300                     & 0.2567 & $\times$ & 0.0181 & 0.2513       & 0.0392 & 0.0000        \\
\multicolumn{1}{l|}{} & S & D3 \cite{gozuaccik2019unsupervised}       & \textit{0.7097} & 0.6829                     & \textbf{0.7445}     & \textbf{0.9471} & \textit{0.7142}       & 0.5613 & 0.4340          \\
\multicolumn{1}{l|}{} & U & NN-DVI \cite{liu2018accumulating}     & 0.6789 & 0.5889 & $\times$ & 0.4550 & 0.4134       & 0.2428 & 0.2327         \\
\multicolumn{1}{l|}{} & SS & ERICS \cite{haug2021learning}     & 0.6371 & 0.6345 & 0.5604 &     0.3572 &
  0.4028 &
  0.4147 &
  0.4312         \\
\multicolumn{1}{l|}{\multirow{-8}{*}{\rotatebox{90}{SoA}}} & U &
  CDLEEDS \cite{DBLP:conf/cikm/HaugBZK22}  &
  0.6556 &
  0.6584 &
  0.5771 &
  \textit{0.7136} &
  0.6625 &
  \textit{0.6625} &
  \textit{{\ul 0.6625}} 
  \\ \midrule
\multicolumn{1}{l}{} & U &
  DynAmo [us] &
  \textbf{0.7159} &
  \textit{0.8187} &
  \textit{{\ul 0.7407}} &
  0.6988 &
  \textbf{0.7601} &
  \textbf{0.7699} &
  \textbf{0.6686} \\ \bottomrule
\end{tabular}%
}
\end{table}

\subsection{Discussion and Ablation Study}\label{sec:discussion}

\textbf{DynAmo is a promising fully unsupervised drift detector for complex and realistic domains:} Table \ref{tab:best_performances} shows the performances in terms of average F1 scores on 30 runs for each dataset. Since the Table does not suggest a clear winner over all datasets, we conducted a Friedman Test on the illustrated average F1 scores. Here, we obtain a test statistic equal to $77.601$ with a p-value of $4.4895\times 10^{-10}$. Since the p-value is less than 0.05, we can reject the null hypothesis that the average F1 scores across all datasets are the same for all methods. To verify whether DynAmo has statistically and significantly different (better) average F1 scores across the board, we perform a Bonferroni-Dunn post-hoc test where the control detector is DynAmo. The test suggests that DynAmo is statistically and significantly different across the board, across the board, to Pelt, BinSeg, Window, BottomUp, IKSSW, IKSS-bdd, MD3-EGM, STUDD, and KLCPD.
The average F1 scores per dataset suggest that DynAmo is the best detector on $4/7$ datasets, underperforms in $2/7$ cases (while keeping the second and third position), and performs equally well on $1/7$ scenarios according to a one-way ANOVA with post-hoc Tukey HSD (p-value of 0.05). Overall, DynAmo is the best-performing method, despite it being unsupervised, surpassing the supervised D3 by $7.91\%$ on average on all datasets.

We also observe that DynAmo has very stable performance across all datasets, a property shared only with some of the baseline methods - which have far fewer parameters. This desirable feature depends on the ensemble-based predictive strategy that ensures stability in different contexts.

Regarding the two types of datasets, we note that performances on the synthetic datasets are better on average for all systems due to the less challenging scenarios  (see Sec. \ref{sec:dataset}). Here, Dynamo is surpassed only in the AS dataset by the supervised MD3 approach. In the challenging and realistic EL datasets, Dynamo always holds the lead except for ELP1-D, where the supervised D3 and the unsupervised CDLEEDS surpass it.
Since ELP1-D represents a patient with homogeneous behavioural patterns (as far as the duration of sleep) throughout the normality period, D3's underlying trained classifier can easily devise a separation hyperplane between anomalous and normal instances. However, D3's performances drop in the other three harder and oscillatory scenarios. We note that all (semi)supervised methods (i.e., D3, MD3, MD3-EGM, STUDD, and ERICS) fail to output a decision (i.e., $F1=0$) where the underlying classifier is not capable of distinguishing between normal and abnormal behaviour patterns.

\begin{figure}[!t]
    \centering
    \resizebox{\linewidth}{10cm}{\includegraphics{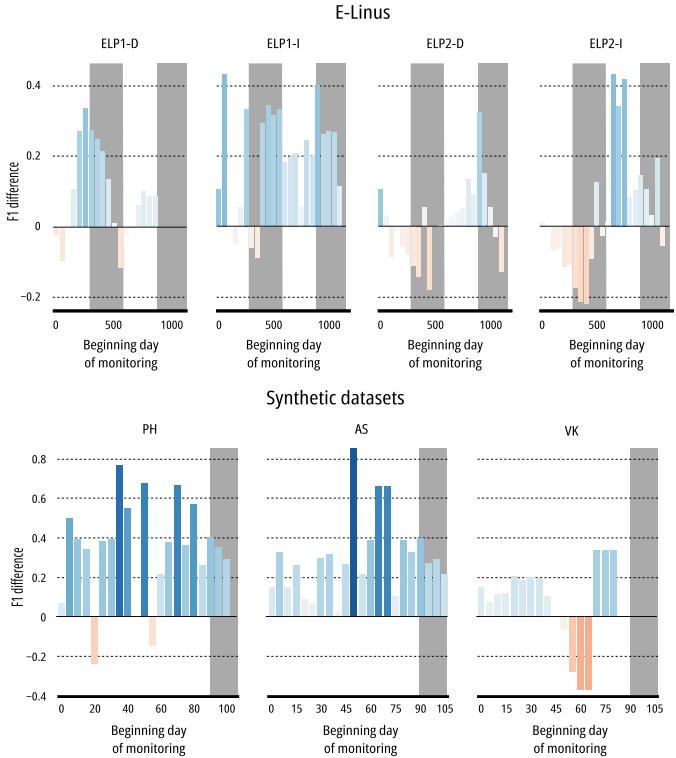}}
    \caption{DynAmo vs CDLEEDS: varying the start offset of the monitoring time (x-axis), the plots depict the difference in terms of F1 scores between DynAmo and CDLEEDS (y-axis). For visualisation purposes we depict the difference in performances according to a red-blue divergent colour scale where blue indicates that DynAmo outperforms CDLEEDS and red vice versa. The shaded area depicts the drift period.}
    \label{fig:el_start_offset}
\end{figure}
\noindent\textbf{DynAmo is robust to distributional changes within the drift period:} During the drift period, an observed behaviour changes until a new "normality" pattern is adopted. We want to demonstrate that DynAmo is robust against any setting of the reference and detection windows throughout the trajectory given in input. Since we are in an unsupervised scenario, we compare DynAmo's performances with CDLEEDS, the second-best unsupervised approach according to the sum of ranks produced by the Friedman test on all datasets\footnote{DynAmo has a sum of ranks equal to $115$, while CDLEEDS $104$; where the highest sum of ranks is $17\times 7 = 119$, i.e., the number of compared methods for all the $7$ datasets.}. In Figure \ref{fig:el_start_offset}, we show the difference in terms of F1 scores between DynAmo and CDLEEDS on patients P1 and P2 of EL. Positive (blue) values indicate that DynAmo outperforms CDLEEDS. Negative (red) values show that CDLEEDS is better than DynAmo. The shaded areas represent the (ground-truth) drift periods. The experiment shows that DynAmo performs better than CDLEEDS in general. It does so also when the monitoring time starts slightly before or within the drift period - the grey area in Figure \ref{fig:el_start_offset})- although in some cases, e.g., for patient P2,  CDLEEDS has the upper hand during the first portion of the drift trajectory. 
For the synthetic datasets, it is interesting to notice how - in line with what is reported in Table \ref{tab:dataset_characteristics} - for VK, both strategies are competitive until the near end of the behavioural sequence. In this case, neither DynAmo nor CDLEEDS can capture anomalous behaviours when the beginning monitoring day is within the drift period. Contrarily, for PH and AS, DynAmo outperforms CDLEEDS.

\begin{figure}[!t]
    \centering
    \includegraphics[width=\linewidth]{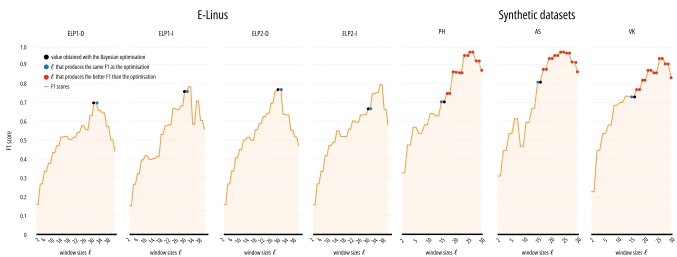}
    \caption{DynAmo's performances in terms of F1 scores when varying the amount of daily hyperboxes (window size) $\ell$.}
    \label{fig:limit_per_window}
\end{figure}

\begin{figure}[!t]
    \centering
    \includegraphics[width=\linewidth]{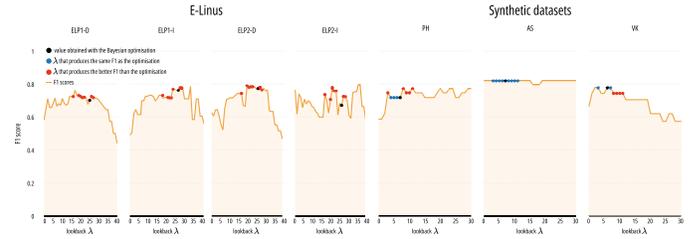}
    \caption{DynAmo's performances in terms of F1 scores when varying the look-back parameter $\lambda$.}
    \label{fig:lookback}
\end{figure}

\noindent\textbf{More daily hyperboxes for testing the distributional shifts imply more confident detections:} Figure \ref{fig:limit_per_window} depicts the contribution of the amount of daily hyperboxes $\ell$ used to populate the two windows. Here, we set $\ell \in [2,40]$ for EL and $\ell \in [2,30]$ for the synthetic datasets and leave the other hyperparameters unchanged. With a black circle, we report the average performances for all datasets reached according to the Bayesian optimisation (see Sec. \ref{sec:compared_methods}). Additionally, a blue circle illustrates similar performances as the ones reported from the optimisation with a different choice of the window size $\ell'$. Red circles represent the choices of the window size, which produce better average results than those reported in Table \ref{tab:best_performances}, leaving the other hyperparameters unchanged. This phenomenon is present specifically in the synthetic datasets, which leads us to believe that the optimisation strategy reached a local minimum, pruning the trials that might have generated a better combination of hyperparameters with higher average F1 scores on the datasets. For the synthetic datasets, the F1 curve keeps increasing due to the homoscedasticity of the "normal" period in the behavioural sequence.

Nevertheless, we need to cope with \textit{real-world critical scenarios where the promptness of prediction is a crucial aspect and not much time is spent in training/updating the models}. In this context, the fewer daily profiles needed to make predictions, the more robust the system is. We invite the reader to notice that $\lfloor \frac{\ell}{2} \rfloor = 15$ days are a reasonable amount of time to build the reference/detection window (i.e., only $\sim1\%$ of the total monitoring days) in the EL  datasets. Similarly, for the synthetic datasets, although we can reach better F1 scores with larger $\ell$ - see the red dots - we argue that coping with the cold start problem and capping the window size is more beneficial. We perform a similar study for MD3, MD3-EGM, D3, and ERICS in Fig. \ref{fig:window_size_change}.

\noindent\textbf{A short/medium-term look-back has benefits in dealing with recurrent normal behaviour:} Figure \ref{fig:lookback} illustrates the contribution of the look-back amount $\lambda$ used to trace the evolution of the feature hyperboxes within the same window. The circles have the same meaning as presented in the previous ablation study (see Figure \ref{fig:limit_per_window}), now with $\lambda$ as the parameter of interest. Notice that a large $\lambda$ in EL degrades the performances in almost all cases besides ELP2-I. In particular, we argue that maintaining the history of $\lambda$ days before the current might help have a complete overview of the hyperbox evolution in time. However, it does not always represent the current behavioural situation due to outdated routine activities. ELP2-I presents a sawtooth-like trend, which leads us to believe that the noise in the daily routines does not permit DynAmo to build an effective evolutionary view of the feature hyperboxes in time (see Sec. \ref{sec:datasets_characteristics}).

Additionally, we notice that a $\lambda \in [1,30]$ has a substantial performance gain w.r.t. $\lambda=0$ because daily routines tend to re-occur according to a specific seasonality trend. More specifically, $\lambda$ allows DynAmo to learn this intrinsic and latent seasonality in the evolution of the feature hyperboxes (e.g., sleeping less in hot seasons). Contrarily, the synthetic datasets do not have a clear shared monotonicity of the F1 scores when varying $\lambda$. Moreover, one can notice how looking back for too much leads to worse performances than not looking back at all. For this reason, a trade-off between $\lambda$ (past), $\ell$ (current and future), and the trajectory length is necessary for real-world complex scenarios. 

\section{Conclusion}\label{sec:conclusion}

We presented a drift anomaly detection framework for multivariate symbolic sequences, such as human behavioural patterns. Our approach, DynAmo, is based on dynamically clustering the monitored events of the same type with a selected frequency (e.g., days or weeks), generating a trajectory by extracting features from the densest clusters of centroids. Finally, DynAmo exploits an ensemble of trackers/divergence tests to predict a drift on the reference or detection windows. 

In summary, DynAmo has the following notable features:
\begin{enumerate}
    \item It is fully unsupervised, which is desirable, particularly for personalised learning applications and in contexts where it is not easy to obtain labelled data.
 
    \item Dynamo surpasses, on average, supervised, unsupervised and semi-supervised systems with a stronger advantage in complex, realistic scenarios (such as those of the E-Linus patients telemonitoring dataset).

    \item Contrary to other compared systems, DynAmo is robust to oscillatory input signals, thanks to the look-back $\lambda$ parameter. Furthermore, the look-back strategy allows for detecting drift anomalies even when they occur during the construction of the reference window. This is a relevant property since we cannot guarantee that drifts do not occur during the initial observation period.

    \item Performances are stable across datasets due to the prediction strategy based on an ensemble of trackers and divergence tests.
\end{enumerate}

A limitation of DynAmo is that it needs to integrate point and drift anomalies into a single detection framework. However, it can identify potential outliers as low-density peripheral clusters (see Sec. \ref{sec:dynamic_clustering}). DynAmo also lacks explainability mechanisms, for example, to detect causal patterns among detected anomalies. Finally, we have omitted some technical details on how drifting behaviours are collected and analyzed in applications. For example,  in some application scenarios, it may be relevant to recognize that a detected anomaly has occurred previously or is recurring (see Sec. \ref{sec:cyclic_anomaly_detection}). We leave these extensions to future studies.

\section*{Acknowledgment}
This work has been partly supported by the project  “E-DAI”: Digital ecosystem for integrated analysis of heterogeneous health data relating to high-impact pathologies: an innovative model of assistance and research. Piano Operativo Salute (POS) 2014-2020, CUP: B83C22004150001

\ifCLASSOPTIONcaptionsoff
  \newpage
\fi

\bibliographystyle{IEEEtran}
\bibliography{bibliography}

\begin{IEEEbiographynophoto}{\href{https://scholar.google.com/citations?user=JIidltYAAAAJ}{Bardh Prenkaj}} obtained his M.Sc. and PhD in Computer Science from the Sapienza University of Rome in 2018 and 2022, respectively. He then worked as a senior researcher at the Department of Computer Science in Sapienza. He led a team of four junior researchers and six software engineers in devising novel anomaly detection strategies for social isolation disorders. Since October 2022, he has been a postdoc at Sapienza in eXplainable AI and Anomaly Detection. He is part of and actively collaborates with Italian and international research groups (\href{https://www.gov.sot.tum.de/rds/overview/}{RDS} of TUM, \href{https://www.pinlab.org/}{PINlab} of Sapienza,  and \href{https://cs.gmu.edu/~dmml/}{DMML} of GMU). He serves as a Program Committee (PC) member for conferences such as ICCV, CVPR, KDD, CIKM, and ECAI. He actively contributes as a reviewer for journals, including TKDE, TKDD, VLDB, TIST, and KAIS.
\end{IEEEbiographynophoto}

\begin{IEEEbiographynophoto}{\href{https://scholar.google.com/citations?user=yf0g6zkAAAAJ&hl=en&oi=ao}{Paola Velardi}} is a full professor with the Department of Computer Science at Sapienza University of Rome. She has been working on Artificial Intelligence since 1983, when she was a visiting researcher at Stanford University. Her main research interests are in the areas of text processing, machine learning, and knowledge-bases.  Applications of interest are in the areas of social networks, e-health,  e-learning, bioinformatics, recommender systems, cultural heritage, machine learning for visual analytics. She published over 200 papers in all major scientific journals and venues. She has also designed and coordinated many projects dedicated to bridging the gender gap in ICT.
\end{IEEEbiographynophoto}

\newpage

\begin{appendices}

\section{Feature processing and event filtering}\label{sec:feature_extraction}

\begin{table}[!h]
\centering
\caption{The description of the cumulative functions $Z_1,\dots,Z_5$. Notice that we use the first-order logic to describe the meaning of the features extracted from the set of daily events}
\label{tab:features_extracted}
\resizebox{\linewidth}{!}{%
\begin{tabular}{@{}llll@{}}
\toprule
Cumulative function &
  Feature extracted &
  Description &
   \\ \midrule
$Z_1(\mathcal{E}_j)$ &
  $\underset{b(e_i)}{\arg \min}\{e_i \in \mathcal{E}_j\}$ &
  \begin{tabular}[c]{@{}l@{}}The beginning time of the first\\ sleep event in the $j$-th $\Delta$ interval\end{tabular} &
   \\
   \midrule
$Z_2(\mathcal{E}_j)$ &
  $\underset{f(e_i)}{\arg \max}\{ e_i \in \mathcal{E}_j\}$ &
  \begin{tabular}[c]{@{}l@{}}The end time of the last sleep\\ event in the $j$-th $\Delta$ interval\end{tabular} &
   \\
   \midrule
$Z_3(\mathcal{E}_j)$ &
  $\underset{{e_i \in \mathcal{E}_j}}{\sum} \big(f(e_i) - b(e_i)\big)$ &
  The cumulative daily sleep portion &
   \\
   \midrule
$Z_4(\mathcal{E}_j)$ &
  $|\mathcal{E}_j|$ &
  \begin{tabular}[c]{@{}l@{}}The number of distinct sleep\\ events depicts the number of \\interruptions in the $j$-th $\Delta$ interval\end{tabular} &
   \\
   \midrule
$Z_5(\mathcal{E}_j)$ &
  \begin{tabular}[c]{@{}l@{}}

$\underset{\substack{e'_i,e''_i \in \mathcal{E}_j\\\land\;\text{o}(e''_i) = \text{o}(e'_i)+1}}{\sum}{\big(b(e''_i) - f(e'_i)\big)}$      \\ where $\text{o}(e)$ is the order in which\\the events re sorted in $\mathcal{E}_j$\end{tabular} 
&
  \begin{tabular}[c]{@{}l@{}}The cumulative duration of sleep\\ interruptions in the $j$-th $\Delta$ interval\end{tabular} &
   \\ \bottomrule
\end{tabular}%
}
\end{table}

In the E-Linus (EL) and the synthetic datasets, we consider the anomalies over trajectories of nocturnal sleep events. It is important to analyse sleep patterns because many disorders can affect it \cite{sleep}, becoming chronic over time. 

To effectively represent the daily sleep behaviour of a particular patient, we begin our daily monitoring between 21:00 of the day $j$ and noon of $j+1$. Notice that $\mathbb{X}_j \in \mathbb{R}^5$ now is a vector of five dimensions since we extract five different features from the events in the $j$-th $\Delta$ interval. For convenience purposes, we denote with $\mathcal{E}_j$ the set of events collocated within the temporal bounds of the $j$-th $\Delta$ interval.
\[
    \mathbb{X}_j = \begin{bmatrix}
        Z_1(\mathcal{E}_j) &
        Z_2(\mathcal{E}_j)&
        Z_3(\mathcal{E}_j)&
        Z_4(\mathcal{E}_j)&
        Z_5(\mathcal{E}_j)
    \end{bmatrix}
\]
Table \ref{tab:features_extracted} illustrates the features extracted and their description for each cumulative function $Z$.

\section{Dataset characteristics}\label{sec:datasets_characteristics}

\begin{figure*}
    \centering
    \includegraphics[width=\textwidth]{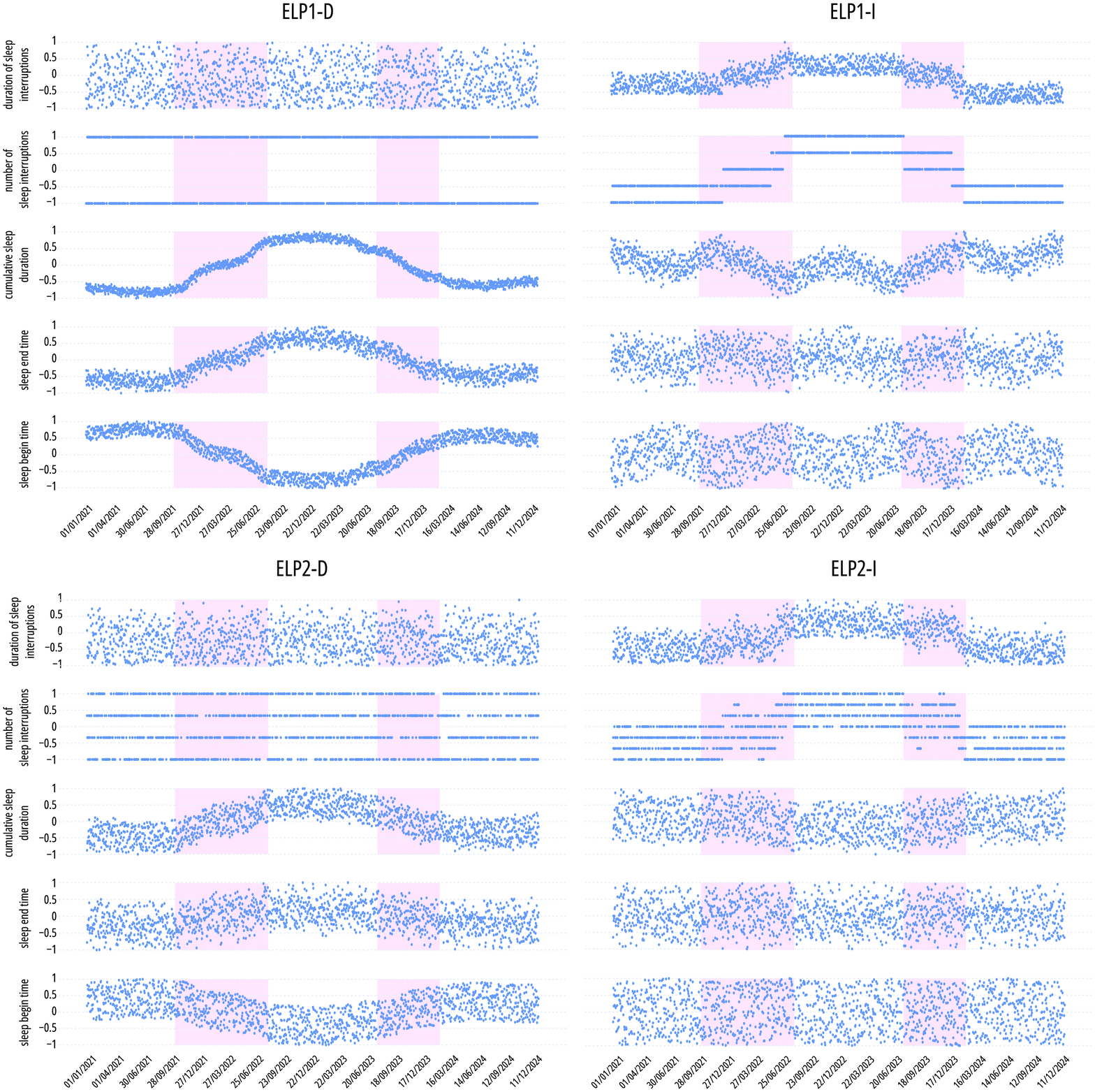}
    \caption{The five features described in Section \ref{sec:feature_extraction} normalised in the range [-1,1] for each patient in E-Linus. The highlighted portions of the trajectory illustrate the drift period.}
    \label{fig:EL_dataset_features}
\end{figure*}

\begin{figure*}
    \centering
    \includegraphics[width=\textwidth]{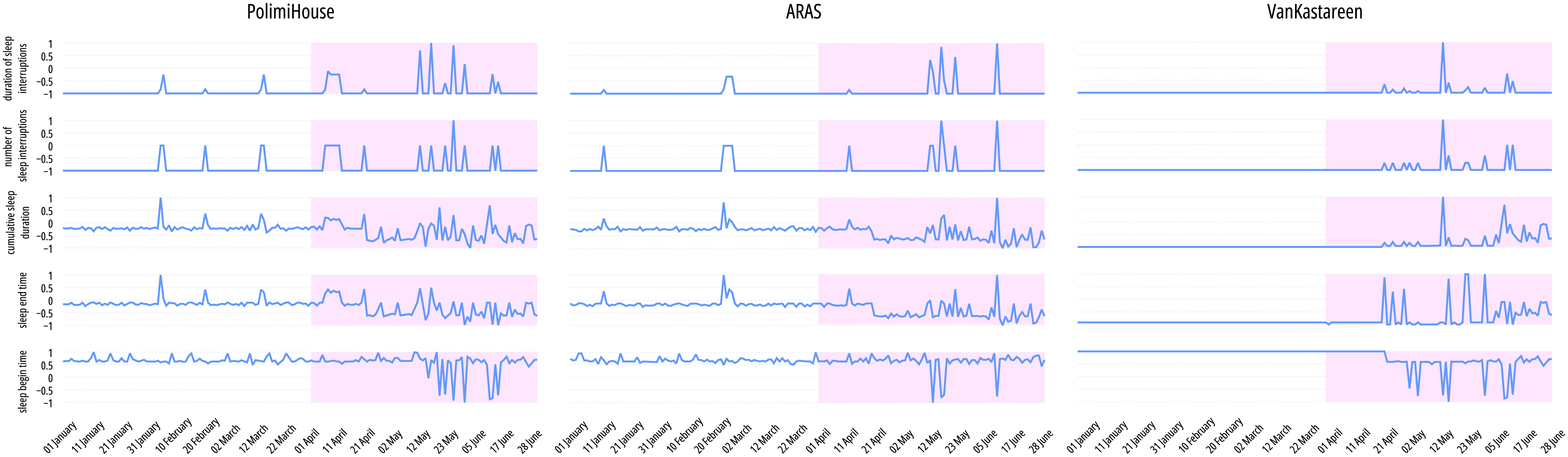}
    \caption{The five features described in Sec. \ref{sec:feature_extraction} normalised in the range [-1,1] for the synthetic datasets. The highlighted portions of the trajectory illustrate the drift period. Notice how, differently from EL, the normal behavior of the patient is mostly homoscedastic with rare distributional outliers.}
    \label{fig:synthetic_dataset_features}
\end{figure*}

Here, we present the characteristics of the datasets examined in the main manuscript to make it easier for the reader to follow the discussion of Section 5.3. In Figures \ref{fig:EL_dataset_features} and \ref{fig:synthetic_dataset_features}, we illustrate the trend of each feature described in Table \ref{tab:features_extracted} for, respectively, EL and the synthetic datasets. For visualization purposes, we depict scatter plots in the scenario of EL due to the very long length of this dataset\footnote{Remember however that we started from a much shorter monitoring period, six weeks, that we artificially but realistically prolonged and described in Sec. 4 of the main paper.} (4 years). Instead, we opt for line charts for the synthetic datasets to emphasize the artificial trend of the patient trajectories. Besides depicting real/realistic patient behaviours, notice how not all features in EL have a clear trend. When the duration of the sleep patterns is perturbed in the first patient (ELP1-D), the third feature - i.e. cumulative daily duration of sleep - has a clear drifting trend, whereas the features corresponding to sleep interruptions - see the first two subplots - have a completely irregular pattern which hinders DynAmo and SOTA detectors from performing well. Contrarily, when leaving the sleep duration untouched, in ELP2-I, there is no evident trend in the last two subplots, while the number of interruptions - see the second subplot - delineates a clear shift in distributions within the drift periods. Considering more heterogeneous behavioural patterns in the second patient, the detection becomes even more arduous, as reported in the ablation study of the performances of DynAmo and CDLEEDS when offsetting the beginning of the monitoring time (see the plots for ELP2-D and ELP2-I).

In the synthetic datasets, notice how the patient "normality" is simpler than that in the real-world scenarios in EL. Particularly, in VanKastareen (VK), the normality does not aid methods that are based on the evolution of the feature space - e.g. DynAmo, CDLEEDS \cite{DBLP:conf/cikm/HaugBZK22}, and ERICS \cite{haug2021learning} - because of its homoscedasticity. Moreover, notice how, in the drift period, the shift in distribution is not gradual, as illustrated in EL. This rapid change in distribution makes it difficult for the hyperbox evolution to grasp any significant trend. Exploiting a window size of 7 (i.e. $\lfloor \frac{\ell}{2} \rfloor = \frac{14}{2}$), DynAmo is incapable of capturing any clear evolution when the minimum and maximum span of the feature space remains unvaried (or slightly varies), especially when the beginning monitoring time shifts towards the end of the trajectory. We suspect that reconstruction-based methods would perform well in the synthetic dataset scenarios due to the homogeneity of the normal period, which would aid the underlying model in differentiating the abnormal period more easily. Nevertheless, testing these methods is out of the scope of this work.

\section{Hyperparameter analysis for DynAmo}\label{sec:dynamo_hyperparamters}

\begin{figure}[!t]
    \centering
    \includegraphics[width=\linewidth]{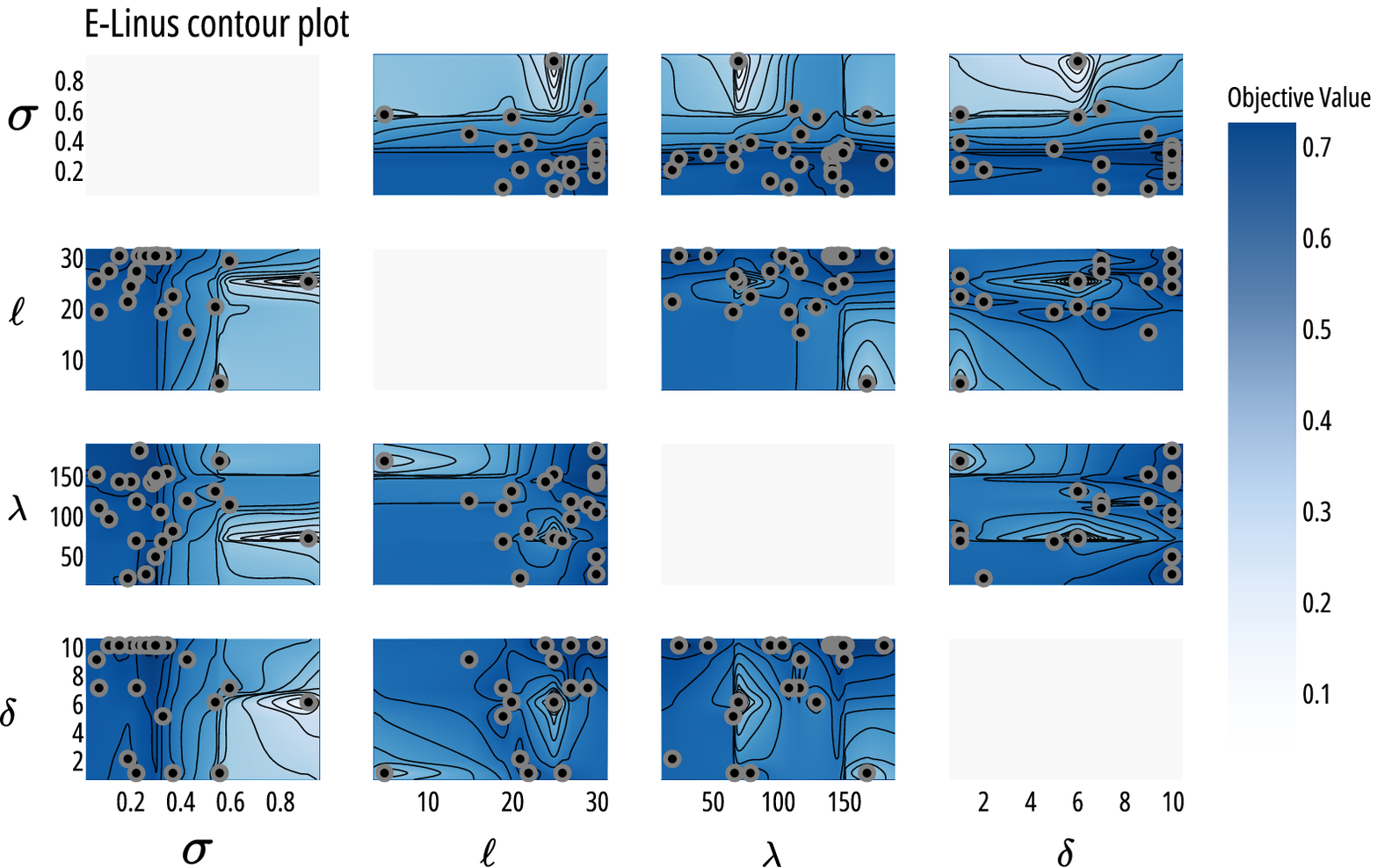}
    \caption{The hyperparameter contour plot in the EL scenario.}
    \label{fig:el_contour}
\end{figure}

\begin{figure}[!t]
    \centering
    \includegraphics[width=\linewidth]{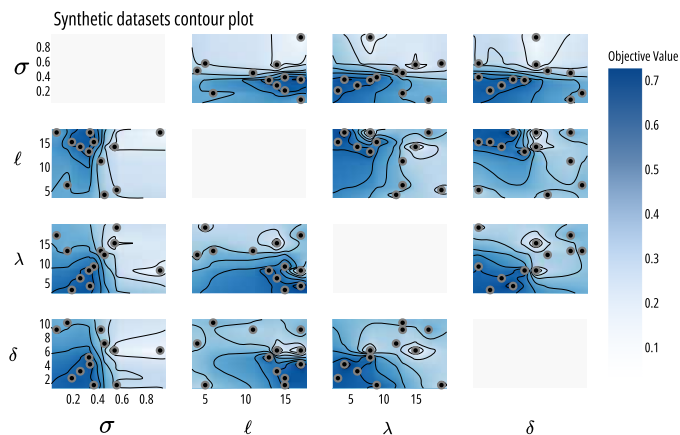}
    \caption{The hyperparameter contour plot in the synthetic dataset scenarios.}
    \label{fig:synthetic_contour}
    \vspace{-.5cm}
\end{figure}

The parameters described in Section 5.2 of the main material are a derivation of a Bayesian optimisation for 100 trials with the following search space for each hyperparameter and dataset:
\begin{itemize}
    \item \textit{EL} - $\lambda \in [1,180]$, $\ell \in [4,30]$, $\delta \in [1,10]$, and $\sigma \in \unif_{0.05}^{0.95}$;
    \item \textit{PH, AS, VK} - $\lambda \in [1,20]$, $\ell \in [4,20]$, $\delta \in [1,10]$, and $\sigma \in \unif_{0.05}^{0.95}$;
\end{itemize}
where $\unif_{x}^{y}$ is a uniform distribution with the low-end equal to $x$ and the high-end to $y$. Instead of performing an optimisation for each dataset, we optimise the choice of the hyperparameters for all datasets belonging to the same group (i.e. real vs synthetic) where the objective function is the average F1 score.

Figures \ref{fig:el_contour} and \ref{fig:synthetic_contour} represent contour plots of the hyperparameter optimisation for EL and the synthetic datasets. Notice that, for each pair of the hyperparameters\footnote{The contour plots for the pairs of two identical hyperparameters - e.g., $(\sigma,\sigma)$ - are not shown because they do not convey any information.}, we illustrate the reached F1 score - see the dots in the subplots - over the objective value region depicted in a light-to-dark colour scale. The darker the region, the nearer the F1 score to the maximally reached value in the optimisation. It is interesting to notice that, throughout the optimisation, DynAmo can reach local minima - see the ridges in the plots - that make the combination of hyperparameters harder to reach the objective value. Furthermore, optimisation in EL is concentrated on the darker region sooner than in synthetic datasets, where the F1 scores are distributed heterogeneously throughout the regions.

\begin{figure}
    \centering
    \includegraphics[width=\linewidth]{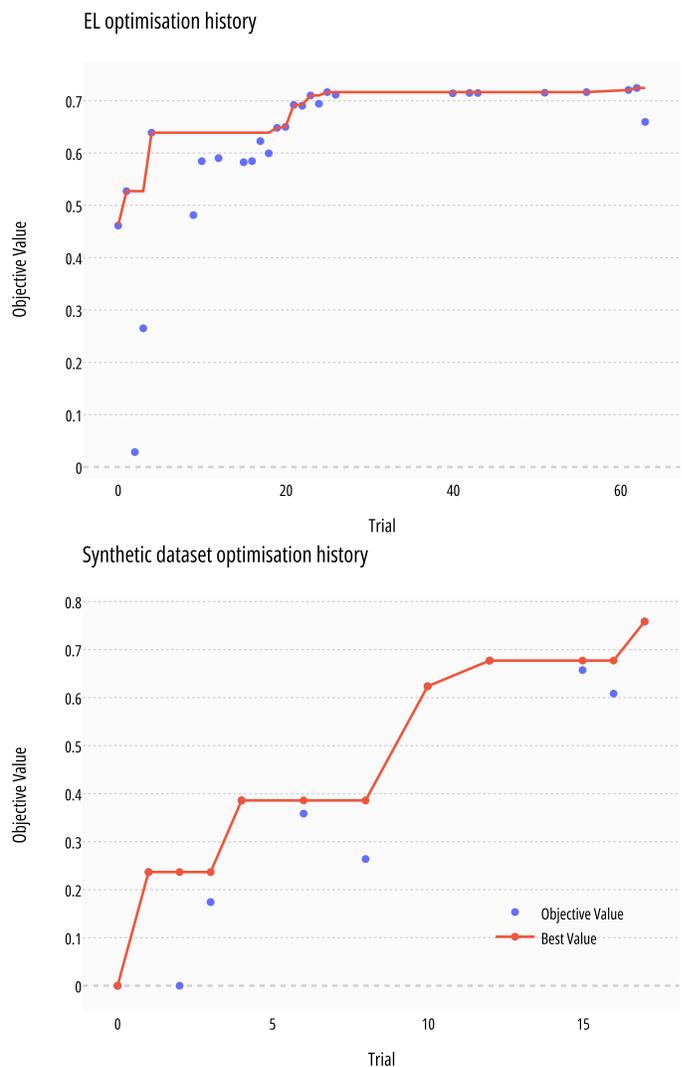}
    \caption{The optimisation history for EL and the synthetic datasets.}
    \label{fig:opt_history}
\end{figure}

Figure \ref{fig:opt_history} depicts the optimisation history of EL and the synthetic datasets according to the combination of the hyperparameters described above. The x-axis depicts the trial, and the y-axis the objective value (i.e., the F1 score). Notice that the maximum number of trials set in the optimisation is 100. Nevertheless, the Bayesian optimiser can prune the current trial if its difference in the objective value does not exceed a certain threshold. Therefore, the x-axis in the two plots in the Figure is cut off. As Figure \ref{fig:el_contour} shows, the objective value converges to the best obtainable value curve (see the concentration of the blue dots w.r.t. the red curve).

Meanwhile, the optimisation for the synthetic datasets follows a more spurious trend even though the final objective value is higher than that in EL. Finally, if one compares Figure \ref{fig:synthetic_contour} with the plot on the right in Figure \ref{fig:opt_history}, the Bayesian optimisation behaves like a random optimisation strategy. This leads us to believe that the absence of anomalies in the synthetic datasets' sequences before the monitoring time's end does not make DynAmo adapt to potential false positives/negatives when detecting drifts.

\begin{figure*}
    \centering
    \includegraphics[width=\textwidth]{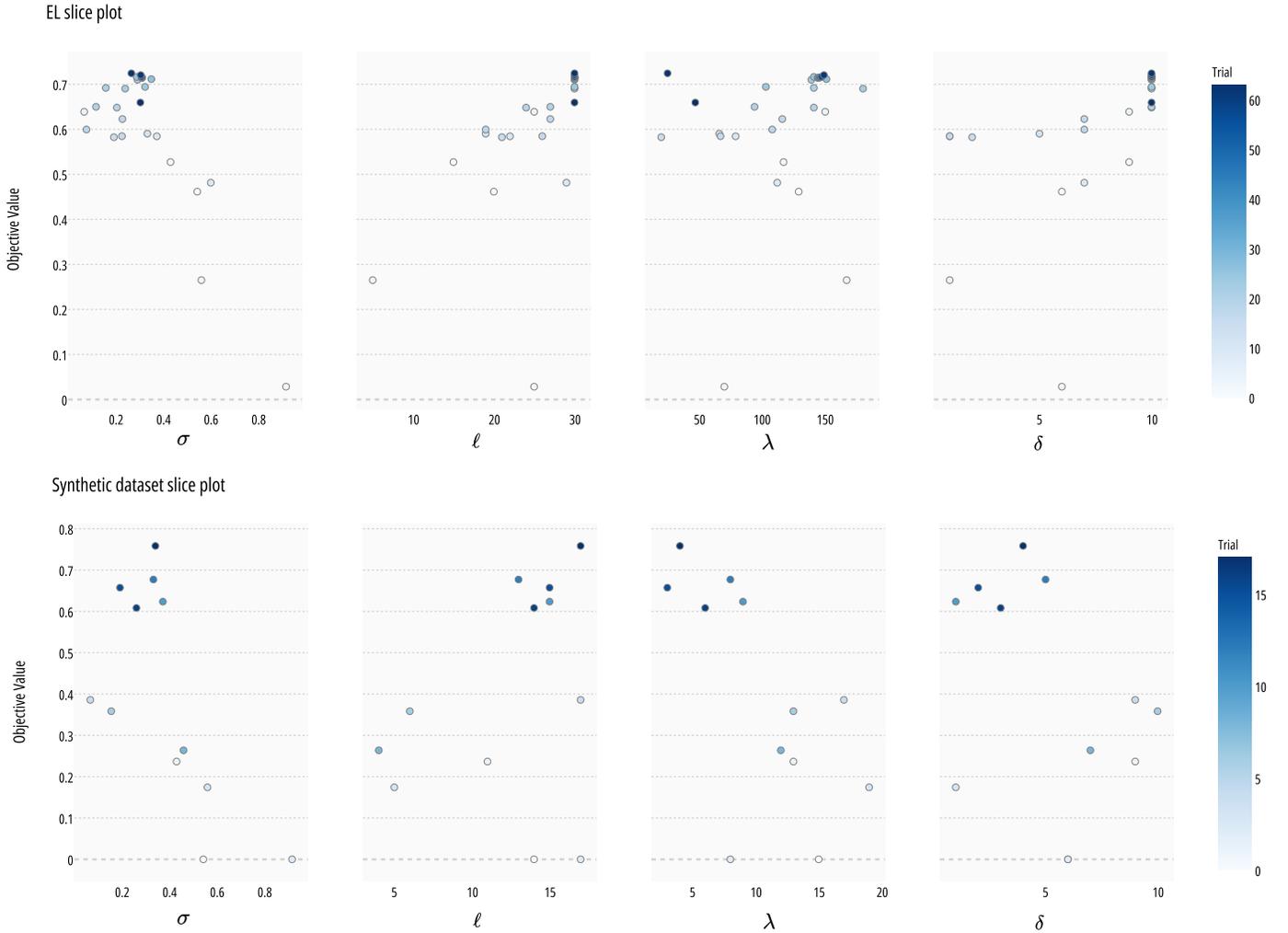}
    \caption{The objective value reached for each hyperparameter through the optimisation trials.}
    \label{fig:slice_plots}
\end{figure*}

For completeness, we show the slice plots (see Figure \ref{fig:slice_plots}) for all the hyperparameters contributing to the objective value according to the number of trials passed in the optimisation process.

\section{Compared methods and hyperparameter optimisation}\label{sec:compared_methods}

\begin{table*}[!t]
\centering
\caption{Search space of the hyperparameters of the baselines.}
\label{tab:baselines_search_space}
\resizebox{\textwidth}{!}{%
\begin{tabular}{@{}llllllll@{}}
\toprule
  &
  BinSeg &
  BottomUp &
  PELT &
  Window &
  IKSSW &
  IKSS-bdd &
  KernelCPD \\ \midrule
  EL &
  \begin{tabular}[t]{@{}l@{}}$\ell \in [4,30]$\\ $\text{pen} \in [1,10]$\\ $\text{min\_size} \in [2,10]$\\ $\text{jump} \in [5,10]$\\ $\text{model} \in \{\text{l1},\text{l2},\text{rbf}\}$\end{tabular} &
  \begin{tabular}[t]{@{}l@{}}$\ell \in [4,30]$\\ $\text{pen} \in [1,10]$\\ $\text{min\_size} \in [2,10]$\\ $\text{jump} \in [5,10]$\\ $\text{model} \in \{\text{l1},\text{l2},\text{rbf}\}$\end{tabular} &
  \begin{tabular}[t]{@{}l@{}}$\ell \in [4,30]$\\ $\text{pen} \in [1,10]$\\ $\text{min\_size} \in [2,10]$\\ $\text{jump} \in [5,10]$\\ $\text{model} \in \{\text{l1},\text{l2},\text{rbf}\}$\end{tabular} &
  \begin{tabular}[t]{@{}l@{}}$\text{width} \in [2,10]$\\ $\text{pen} \in [1,10]$\\ $\text{min\_size} \in [2,10]$\\ $\text{jump} \in [5,10]$\\ $\text{model} \in \{\text{l1},\text{l2},\text{rbf}\}$\end{tabular} &
  \begin{tabular}[t]{@{}l@{}}$\ell \in [4,30]$\\ $\delta \in [1,10]$\end{tabular} &
  \begin{tabular}[t]{@{}l@{}}$\ell \in [4,30]$\\ $\delta \in [1,10]$\end{tabular} &
  \begin{tabular}[t]{@{}l@{}}$\ell \in [4,30]$\\ $\text{pen} \in [1,10]$\\ $\text{min\_size} \in [2,10]$\\ $\text{kernel} \in \{\text{linear},\text{cosin},\text{rbf}\}$\end{tabular} \\ \midrule
  Synthetic &
  \begin{tabular}[t]{@{}l@{}}$\ell \in [4,20]$\\ $\text{pen} \in [1,10]$\\ $\text{min\_size} \in [2,10]$\\ $\text{jump} \in [5,10]$\\ $\text{model} \in \{\text{l1},\text{l2},\text{rbf}\}$\end{tabular} &
  \begin{tabular}[t]{@{}l@{}}$\ell \in [4,20]$\\ $\text{pen} \in [1,10]$\\ $\text{min\_size} \in [2,10]$\\ $\text{jump} \in [5,10]$\\ $\text{model} \in \{\text{l1},\text{l2},\text{rbf}\}$\end{tabular} &
  \begin{tabular}[t]{@{}l@{}}$\ell \in [4,20]$\\ $\text{pen} \in [1,10]$\\ $\text{min\_size} \in [2,10]$\\ $\text{jump} \in [5,10]$\\ $\text{model} \in \{\text{l1},\text{l2},\text{rbf}\}$\end{tabular} &
  \begin{tabular}[t]{@{}l@{}}$\text{width} \in [2,10]$\\ $\text{pen} \in [1,10]$\\ $\text{min\_size} \in [2,10]$\\ $\text{jump} \in [5,10]$\\ $\text{model} \in \{\text{l1},\text{l2},\text{rbf}\}$\end{tabular} &
  \begin{tabular}[t]{@{}l@{}}$\ell \in [4,20]$\\ $\delta \in [1,10]$\end{tabular} &
  \begin{tabular}[t]{@{}l@{}}$\ell \in [4,20]$\\ $\delta \in [1,10]$\end{tabular} &
  \begin{tabular}[t]{@{}l@{}}$\ell \in [4,20]$\\ $\text{pen} \in [1,10]$\\ $\text{min\_size} \in [2,10]$\\ $\text{kernel} \in \{\text{linear},\text{cosin},\text{rbf}\}$\end{tabular} \\ 
  \bottomrule
\end{tabular}%
}
\end{table*}

\begin{table*}[!t]
\centering
\caption{Search space of the hyperparameters of the SoA.}
\label{tab:soa_search_space}
\resizebox{\textwidth}{!}{%
\begin{tabular}{lllllllll}
\toprule
&
  KLCPD &
  MD3 &
  MD3-EDM &
  STUDD &
  D3 &
  NN-DVI &
  ERICS &
  CDLEEDS \\ \midrule
  EL &
  \begin{tabular}[t]{@{}l@{}}$\ell \in [4,30]$\\ $\text{trn\_ratio} \in [0.3,0.5]$\\ $\text{val\_ratio} \in [0.3,0.5]$\\ $\text{max\_iter} \in [5,50]$\\ $\text{batch\_size} \in \{1,2,4,8,16,32,64\}$\\ $\lambda_{\text{ae}} \in [1e-5, 1]$\\ $\lambda_{\text{real}} \in [1e-5, 1]$\\ $\text{eval\_freq} \in [5,25]$\\ $\text{weight\_clip} \in [10^{-2},10^{-1}]$\\ $\sigma \in [0.01,0.95]$\\ $\text{rnn\_hidden\_dim} \in [5,15]$\\ $\text{sub\_dim} \in [1,5]$\\ $\text{optim} \in \{\text{rmsprop},\text{adam},\text{sgd}\}$\\ $\text{lr} \in [1e-4,1e-1]$\\ $\text{weight\_decay} \in [10^{-5},10^{-2}]$\\ $\text{momentum} \in [10^{-2}, 1]$\\ $\text{grad\_clip} \in [5,15]$\\ $\text{critic\_iters} \in [1,10]$\\ \\ \textit{Constraints}:\\ $\text{val\_ratio} > \text{trn\_ratio}$\end{tabular} &
  \begin{tabular}[t]{@{}l@{}}$\sigma \in [0.05, 0.95]$\\ $\ell \in [4,30]$\\ $\text{normal\_history} \in [280,320]$\\ $\text{gamma} \in \{\text{scale},\text{auto}\}$\\ $\text{tol} \in [10^{-4},10^{-3}]$\\ $\text{C} \in [1,10]$\end{tabular} &
  \begin{tabular}[t]{@{}l@{}}$\ell \in [4,30]$\\ $\text{normal\_history} \in [300,320]$\\ $\text{gamma} \in \{\text{scale},\text{auto}\}$\\ $\text{tol} \in [10^{-4},10^{-3}]$\\ $\text{C} \in [1,10]$\\ $\text{sensitivity} \in [10^{-8}, 10^{-2}]$\end{tabular} &
  \begin{tabular}[t]{@{}l@{}}$\text{window\_size} \in [300,320]$\\ $\text{delta} \in [10^{-3}, 0.5]$\end{tabular} &
  \begin{tabular}[t]{@{}l@{}}$\text{w} \in [4,30]$\\ $\text{rho} \in [0.1,2]$\\ $\text{auc} \in [0.75, 0.99]$\end{tabular} &
  \begin{tabular}[t]{@{}l@{}}$\text{batch\_size} \in [2,15]$\\ $\text{k\_nn} \in [4,50]$\\ $\text{sampling\_time} \in [250,750]$\\ $\text{alpha} \in [10^{-3},1]$\\\\\textit{Constraints}:\\$\text{batch\_size} > \text{k\_nn}$\\$\text{sampling\_time} > \text{k\_nn}$\end{tabular} 
  &
  \begin{tabular}[t]{@{}l@{}}$\text{wnd\_mvg\_avg} \in [4,30]$\\ $\text{wnd\_drift\_det} \in [4,30]$\\ $\text{beta} \in [10^{-5}, 10^{-3}]$\\ $\text{init\_mu} \in [0,1]$\\ $\text{init\_sigma} \in [0.1,0.5]$\\ $\text{epochs} = [10,50]$\\ $\text{lr\_mu} \in [10^{-2},10^{-1}]$\\ $\text{lr\_sigma} \in [10^{-3}, 10^{-2}]$\end{tabular} 
  &
  \begin{tabular}[t]{@{}l@{}}$\text{significance} \in [10^{-3}, 10^{-2}]$\\ $\text{gamma} \in [0.05, 0.95]$\\ $\text{max\_node\_size} \in [280,320]$\\ $\text{max\_tree\_depth} \in [1,10]$\\ $\text{max\_time\_stationary} \in \{5,10,25,50,75,100\}$\\ $\text{trn\_ratio} \in [0.05, 0.2]$\\ $\text{baseline\_weight} \in [0.05, 0.95]$\end{tabular} \\ \midrule
  Synthetic 
  & \begin{tabular}[t]{@{}l@{}}$\ell \in [4,20]$\\ $\text{trn\_ratio} \in [0.2,0.3]$\\ $\text{val\_ratio} \in [0.2,0.3]$\\ $\text{max\_iter} \in [5,50]$\\ $\text{batch\_size} \in \{1,2,4,8,16,32,64\}$\\ $\lambda_{\text{ae}} \in [10^{-5}, 1]$\\ $\lambda_{\text{real}} \in [10^{-5}, 1]$\\ $\text{eval\_freq} \in [5,25]$\\ $\text{weight\_clip} \in [10^{-2},10^{-1}]$\\ $\sigma \in [0.01,0.95]$\\ $\text{rnn\_hidden\_dim} \in [5,15]$\\ $\text{sub\_dim} \in [1,5]$\\ $\text{optim} \in \{\text{rmsprop},\text{adam},\text{sgd}\}$\\ $\text{lr} \in [1e-4,1e-1]$\\ $\text{weight\_decay} \in [10^{-5},10^{-2}]$\\ $\text{momentum} \in [10^{-2}, 1]$\\ $\text{grad\_clip} \in [5,15]$\\ $\text{critic\_iters} \in [1,10]$\\ \\ \textit{Constraints}:\\ $\text{val\_ratio} > \text{trn\_ratio}$\end{tabular} 
  &
  \begin{tabular}[t]{@{}l@{}}$\sigma \in [0.05, 0.95]$\\ $\ell \in [4,20]$\\ $\text{normal\_history} \in [80,100]$\\ $\text{gamma} \in \{\text{scale},\text{auto}\}$\\ $\text{tol} \in [10^{-4},10^{-3}]$\\ $\text{C} \in [1,10]$\end{tabular} 
  &
  \begin{tabular}[t]{@{}l@{}}$\ell \in [4,20]$\\ $\text{normal\_history} \in [80, 100]$\\ $\text{gamma} \in \{\text{scale},\text{auto}\}$\\ $\text{tol} \in [10^{-4},10^{-3}]$\\ $\text{C} \in [1,10]$\\ $\text{sensitivity} \in [10^{-8}, 10^{-2}]$\end{tabular} 
  &
  \begin{tabular}[t]{@{}l@{}}$\text{window\_size} \in [70,100]$\\ $\text{delta} \in [10^{-3}, 0.5]$\end{tabular} 
  &
  \begin{tabular}[t]{@{}l@{}}$\text{w} \in [4,20]$\\ $\text{rho} \in [0.1,0.5]$\\ $\text{auc} \in [0.75, 0.99]$\end{tabular} &
  \begin{tabular}[t]{@{}l@{}}$\text{batch\_size} \in [2,10]$\\ $\text{k\_nn} \in [4,20]$\\ $\text{sampling\_time} \in [10,25]$\\ $\text{alpha} \in [10^{-3},1]$\\\\\textit{Constraints}:\\$\text{batch\_size} > \text{k\_nn}$\\$\text{sampling\_time} > \text{k\_nn}$\end{tabular} 
  &
  \begin{tabular}[t]{@{}l@{}}$\text{wnd\_mvg\_avg} \in [4,20]$\\ $\text{wnd\_drift\_det} \in [4,20]$\\ $\text{beta} \in [1e-5, 1e-3]$\\ $\text{init\_mu} \in [0,1]$\\ $\text{init\_sigma} \in [0.1,0.5]$\\ $\text{epochs} = [10,50]$\\ $\text{lr\_mu} \in [10^{-2},10^{-1}]$\\ $\text{lr\_sigma} \in [10^{-3}, 10^{-2}]$\end{tabular} 
  &
  \begin{tabular}[t]{@{}l@{}}$\text{significance} \in [10^{-3}, 10^{-2}]$\\ $\text{gamma} \in [0.05, 0.95]$\\ $\text{max\_node\_size} \in [90,100]$\\ $\text{max\_tree\_depth} \in [1,10]$\\ $\text{max\_time\_stationary} \in \{5,10,25,50,75,100\}$\\ $\text{trn\_ratio} \in [0.05, 0.2]$\\ $\text{baseline\_weight} \in [0.05, 0.95]$\end{tabular} \\ \bottomrule
\end{tabular}%
}
\end{table*}

\begin{table*}[!t]
\centering
\caption{Hyperparameters of the baselines that reach the best performances for the considered scenarios.}
\label{tab:baseline_hyperparams}
\resizebox{\textwidth}{!}{%
\begin{tabular}{@{}llllllll@{}}
\toprule
& BinSeg & BottomUp & PELT & Window & IKSSW & IKSS-bdd & KernelCPD \\ \midrule
 EL 
 & \begin{tabular}[t]{@{}l@{}}$\ell=20$\\ $\text{pen}=4$\\ $\text{min\_size}=3$\\ $\text{jump}=7$\\ $\text{model}=\text{l1}$\end{tabular}
 & \begin{tabular}[t]{@{}l@{}}$\ell=16$\\ $\text{pen}=4$\\ $\text{min\_size}=5$\\ $\text{jump}=7$\\ $\text{model}=\text{l1}$\end{tabular}
 & \begin{tabular}[t]{@{}l@{}}$\ell=16$\\ $\text{pen}=4$\\ $\text{min\_size}=5$\\ $\text{jump}=7$\\ $\text{model}=\text{l1}$\end{tabular}
 & \begin{tabular}[t]{@{}l@{}}$\text{width}=5$\\ $\text{pen}=2$\\ $\text{min\_size}=8$\\ $\text{jump}=6$\\ $\text{model}=\text{l1}$\end{tabular}
 & \begin{tabular}[t]{@{}l@{}}$\ell=30$\\ $\delta=2$\end{tabular}
 & \begin{tabular}[t]{@{}l@{}}$\ell=30$\\ $\delta=1$\end{tabular}
 & \begin{tabular}[t]{@{}l@{}}$\ell=30$\\ $\text{pen}=2$\\ $\text{min\_size}=2$\end{tabular}\\ \midrule 
 Synthetic  
 & \begin{tabular}[t]{@{}l@{}}$\ell=14$\\ $\text{pen}=7$\\ $\text{min\_size}=7$\\ $\text{jump}=10$\\ $\text{model}=\text{l2}$\end{tabular} 
 & \begin{tabular}[t]{@{}l@{}}$\ell=12$\\ $\text{pen}=6$\\ $\text{min\_size}=7$\\ $\text{jump}=9$\\ $\text{model}=\text{l2}$\end{tabular} 
 & \begin{tabular}[t]{@{}l@{}}$\ell=16$\\ $\text{pen}=6$\\ $\text{min\_size}=6$\\ $\text{jump}=10$\\ $\text{model}=\text{l2}$\end{tabular} 
 & \begin{tabular}[t]{@{}l@{}}$\text{width}=7$\\ $\text{pen}=2$\\ $\text{min\_size}=8$\\ $\text{jump}=6$\\ $\text{model}=\text{l1}$\end{tabular} 
 & \begin{tabular}[t]{@{}l@{}}$\ell=20$\\ $\delta=4$\end{tabular} 
 & \begin{tabular}[t]{@{}l@{}}$\ell=20$\\ $\delta=5$\end{tabular} 
 & \begin{tabular}[t]{@{}l@{}}$\ell=16$\\ $\text{pen}=7$\\ $\text{min\_size}=2$\\ $\text{kernel} = \text{linear}$\end{tabular} \\ \bottomrule
\end{tabular}%
}
\end{table*}

\begin{table*}[!t]
\centering
\caption{Hyperparameters of the SoA that reach the best performances for the considered scenarios. The strikeout text shows that the hyperparameter, although optimised, is not used (i.e., \textit{momentum} is useless when the optimiser is different from \textit{SGD}).}
\label{tab:soa_hyperparams}
\resizebox{\textwidth}{!}{%
\begin{tabular}{lllllllll}
\toprule
&
  KLCPD &
  MD3 &
  MD3-EDM &
  STUDD &
  D3 &
  NN-DVI &
  ERICS &
  CDLEEDS \\ \midrule
  EL &
  \begin{tabular}[t]{@{}l@{}}$\ell=30$\\ $\text{trn\_ratio} = 0.3204$\\
  $\text{val\_ratio} = 0.3418$\\
  $\text{max\_iter} = 12$\\ $\text{batch\_size} = 64$\\ $\lambda_{\text{ae}} = 0.1382$\\ $\lambda_{\text{real}} = 0.1966$\\ $\text{eval\_freq} = 12$\\ $\text{weight\_clip} = 0.0839$\\ $\sigma = 0.1013$\\ $\text{rnn\_hidden\_dim} = 14$\\ $\text{sub\_dim} = 1$\\ $\text{optim} = \text{rmsprop}$\\ $\text{lr} = 0.0605$\\ $\text{weight\_decay} = 0.0074$\\ \st{$\text{momentum}=0.0488$}\\ $\text{grad\_clip}=7.8281$\\ $\text{critic\_iters} = 2$\end{tabular} &
  \begin{tabular}[t]{@{}l@{}}$\sigma=0.05$\\ $\ell = 4$\\ $\text{normal\_history} = 320$\\ $\text{gamma} = \text{scale}$\\ $\text{tol} = 8.9035 \times 10^{-4}$\\ $\text{C} = 9.9999$\end{tabular} &
  \begin{tabular}[t]{@{}l@{}}$\ell = 28$\\ $\text{normal\_history} = 320$\\ $\text{gamma} = \text{auto}$\\ $\text{tol} = 5.7601 \times 10^{-4}$\\ $\text{C} = 6.1124$\\ $\text{sensitivity} = 9.2560 \times 10^{-3}$\end{tabular} &
  \begin{tabular}[t]{@{}l@{}}$\text{window\_size} = 320$\\ $\text{delta} = 0.4999$\end{tabular} &
  \begin{tabular}[t]{@{}l@{}}$\text{w} = 12$\\ $\text{rho} = 0.7911$\\ $\text{auc} = 0.8868$\end{tabular} &
  \begin{tabular}[t]{@{}l@{}}$\text{batch\_size} = 15$\\ $\text{k\_nn} = 15$\\ $\text{sampling\_time} = 564$\\ $\text{alpha} = 6.3838 \times 10^{-2}$\end{tabular} &
  \begin{tabular}[t]{@{}l@{}}$\text{wnd\_mvg\_avg} = 10$\\ $\text{wnd\_drift\_det} = 20$\\ $\text{beta} = 7.6941 \times 10^{-4}$\\ $\text{init\_mu} = 0.7570$\\ $\text{init\_sigma} = 0.1865$\\ $\text{epochs} = 50$\\ $\text{lr\_mu} = 5.4294 \times 10^{-3}$\\ $\text{lr\_sigma} = 6.8574 \times 10^{-3}$\end{tabular} &
  \begin{tabular}[t]{@{}l@{}}$\text{significance} = 2.2349 \times 10^{-3}$\\ $\text{gamma} = 0.6566$\\ $\text{max\_node\_size} = 300$\\ $\text{max\_tree\_depth} = 7$\\ $\text{max\_time\_stationary} = 75$\\ $\text{trn\_ratio} = 0.1917$\\ $\text{baseline\_weight} = 0.6230$\end{tabular} \\ \midrule 
  Synthetic &
  \begin{tabular}[t]{@{}l@{}}$\ell=14$\\
  $\text{trn\_ratio} = 0.2121$\\
  $\text{val\_ratio} = 0.2364$\\
  $\text{max\_iter} = 12$\\ $\text{batch\_size} = 1$\\ $\lambda_{\text{ae}} = 0.9415$\\ $\lambda_{\text{real}} = 0.5164$\\ $\text{eval\_freq} = 18$\\ $\text{weight\_clip} = 0.0436$\\
  $\sigma = 0.0337$\\ $\text{rnn\_hidden\_dim} = 9$\\ $\text{sub\_dim} = 1$\\ $\text{optim} = \text{adam}$\\ $\text{lr} = 0.0999$\\ $\text{weight\_decay} = 0.0075$\\ \st{$\text{momentum}=0.9728$}\\ $\text{grad\_clip}=13.5545$\\ $\text{critic\_iters} = 2$\end{tabular} &
  \begin{tabular}[t]{@{}l@{}}$\sigma=0.25$\\ $\ell = 4$\\ $\text{normal\_history} = 96$\\ $\text{gamma} = \text{scale}$\\ $\text{tol} = 10^{-3}$\\ $\text{C} = 9.9999$\end{tabular} &
  \begin{tabular}[t]{@{}l@{}}$\ell = 18$\\ $\text{normal\_history} = 100$\\ $\text{gamma} = \text{auto}$\\ $\text{tol} = 5.7601 \times 10^{-4}$\\ $\text{C} = 6.1124$\\ $\text{sensitivity} = 9.2560 \times 10^{-3}$\end{tabular} &
  \begin{tabular}[t]{@{}l@{}}$\text{window\_size} = 98$\\ $\text{delta} = 0.2570$\end{tabular} &
  \begin{tabular}[t]{@{}l@{}}$\text{w} = 15$\\ $\text{rho} = 0.1842$\\ $\text{auc} = 0.7809$\end{tabular} &
  \begin{tabular}[t]{@{}l@{}}$\text{batch\_size} = 15$\\ $\text{k\_nn} = 15$\\ $\text{sampling\_time} = 564$\\ $\text{alpha} = 6.3838 \times 10^{-2}$\end{tabular}
  &
   \begin{tabular}[t]{@{}l@{}}$\text{wnd\_mvg\_avg} = 17$\\ $\text{wnd\_drift\_det} = 12$\\ $\text{beta} = 10^{-3}$\\ $\text{init\_mu} = 0.7670$\\ $\text{init\_sigma} = 0.3318$\\ $\text{epochs} = 33$\\ $\text{lr\_mu} = 10^{-3}$\\ $\text{lr\_sigma} = 3.3538 \times 10^{-3}$\end{tabular} &
  \begin{tabular}[t]{@{}l@{}}$\text{significance} = 2.8058 \times 10^{-3}$\\ $\text{gamma} = 0.5298$\\ $\text{max\_node\_size} = 91$\\ $\text{max\_tree\_depth} = 10$\\ $\text{max\_time\_stationary} = 50$\\ $\text{trn\_ratio} = 0.05$\\ $\text{baseline\_weight} = 0.5476$\end{tabular} \\ \bottomrule 
\end{tabular}%
}
\end{table*}

The following are the baseline strategies that we compare to in the experiments. As done with DynAmo, we use a Bayesian optimisation (where applicable) for 100 trials for all SoA methods. Tables \ref{tab:baselines_search_space} and \ref{tab:soa_search_space} show the search space for each method in EL and the synthetic scenarios. Tables \ref{tab:baseline_hyperparams} and \ref{tab:soa_hyperparams} show the hyperparameters of each compared method that produce the best performance for each scenario. Notice that for BottomUp, PELT, BinSeg, and KernelCPD, we divide the input trajectory into contiguous windows of length $\lfloor\frac{\ell}{2}\rfloor$. Hence, we have an additional hyperparameter on these methods w.r.t. the original hyperparameters.
\begin{itemize}[topsep=0pt,noitemsep]
    \item \textbf{Keep It Simple (KIS)} is a naive baseline which labels every day randomly. In cases where a strategy underperforms w.r.t. KIS, we can state that it has a random behaviour, making it unfeasible in critical scenarios like e-health. KIS does not have hyperparameters.
    
    \item \textbf{BinSeg} \cite{bai1997estimating,fryzlewicz2014wild} is a greedy procedure that segments a series using a sequential approach in two steps. First, it detects one change point in the complete input series. Then, it splits the series around the change point and repeats these steps on the two resulting sub-series.

    \item\textbf{BottomUp} \cite{keogh2001online} sstarts with many change points and successively deletes the less significant ones. First, it divides the series into sub-series regions along a regular grid. Then, it merges contiguous segments according to a similarity measure.
    
    \item\textbf{PELT} \cite{killick2012optimal,Wambui2015ThePO} relies on pruning criteria to eliminate all possible partitions of a time series. The procedure stops when convergence gets met, thus producing an optimal segmentation of the change points.
    
    \item\textbf{Window} \cite{truong2020selective} uses two sliding windows along the trajectory. It compares statistical properties of the sub-series corresponding to each window according to a discrepancy metric. The discrepancy metric measures how (dis)similar the two sub-series are. When the two windows are highly dissimilar, a change point is signalled.
    
    \item For \textbf{IKSSW}, we use an approach similar to Window, where the discrepancy metric is the Kolmogorov-Smirnov (KS) test. We assume that the first window represents the distribution in the KS test, and the second window is the sample. If the KS test shows that the sample does not come from the distribution, we signal a change point. This approach is an extension of \textbf{IKS-bdd} \cite{dos2016fast}, which performs the KS test between a fixed reference window and a sliding detection one. We use a p-value of $0.01$ for the KS test.
    
    \item\textbf{KernelCPD} \cite{arlot2019kernel} maps the original time series onto a reproducing Hilbert space associated with a user-defined kernel function. The method detects change points by finding mean shifts in the mapped signal while minimising a particular function. Because we do not induce the number of change points a priori, the optimisation procedure is described in \cite{killick2012optimal}. 

\end{itemize}
Additionally, we compare with the following state-of-the-art methods:
\begin{itemize}[topsep=0pt,noitemsep]
    \item For \textbf{KLCPD} \cite{DBLP:conf/iclr/ChangLYP19} we set the maximum number of iterations to 500.
    \item \textbf{MD3} \cite{sethi2015don} uses a linear SVM to calculate the marginal density used to detect drifts.
    \item \textbf{MD3-EGM} \cite{sethi2017reliable} extends MD3.
    \item \textbf{STUDD} \cite{cerqueira2022studd} is a semi-supervised approach that requires the first window observed to contain both classes (i.e., drift and no drift) for it to work properly.
    \item \textbf{D3} \cite{gozuaccik2019unsupervised} relies on an underlying supervised classifier to detect drifts. As observed from the code-base\footnote{\url{https://github.com/ogozuacik/d3-discriminative-drift-detector-concept-drift}}, we exploit the Hoeffding Tree \cite{domingos2000mining} classifier with default parameters.
    \item \textbf{NN-DVI} \cite{liu2018accumulating}  involves the integration of reference and test data batches, followed by generating a normalised version of the adjacency matrix obtained from a k-NN search. Subsequently, the distances between data points in the reference and test sections of the combined adjacency matrix are analysed. These changes in distances are then compared to a predetermined threshold value $\alpha$. To determine this threshold, new reference and test sections are randomly sampled, and the resulting distance changes are used to fit a Gaussian distribution through a statistical analysis.
    \item For \textbf{ERICS} \cite{haug2021learning} we use the probit base model, and fine tune the other parameters.
    \item For \textbf{CDLEEDS} \cite{DBLP:conf/cikm/HaugBZK22} we use the exponential weighted moving average baseline, and fine tune the rest of the hyperparameters.
\end{itemize}

\section{SoA comparison on other metrics}\label{sec:soa_comparison_with_other_metrics}
Recall from Sec. 3.5 of the main material that DynAmo might wrongly replace the reference and detection windows, resulting in a false positive drift alarm. We argue that in real-world clinical scenarios, false negative errors should be carefully treated since it is worse than "real anomalous" behavioural shifts that happened and were missed by DynAmo, leading to potential risks for the monitored patients. However, we note that balancing the cost of false positives to manage intervention costs and the risk to the patient's health arising from false negatives may depend on local policies. Nevertheless, according to a fairness principle, we should weigh human health more than intervention costs. Here, we provide Tables \ref{tab:fpr} and \ref{tab:fnr} that illustrate averages of False Positive Rate (FPR) and False Negative Rate (FNR), respectively, for all methods described in Sec. \ref{sec:compared_methods} on the datasets of Sec. \ref{sec:datasets_characteristics}.

\begin{table}[!h]
\centering
\caption{Average FPR scores (over 30 runs) of SoA methods against DynAmo. Notice that $\times$ represents no convergence. A value is in bold if it is the lowest value on average; it is italic if it is the second best. S denotes supervised learning, SS semi-supervised learning, U unsupervised learning.}
\label{tab:fpr}
\resizebox{\linewidth}{!}{%
\begin{tabular}{@{}lclccccccc@{}}
\toprule
\multicolumn{3}{l}{}              & \multicolumn{3}{c}{Synthetic datasets}           & \multicolumn{4}{c}{Realistic datasets}                        \\ \cmidrule(lr){3-9}
\multicolumn{3}{l}{\multirow{-2}{*}{}} &
  PH &
  AS &
  VK &
  ELP1-D &
  ELP1-I &
  ELP2-D &
  ELP2-I \\ \midrule
\multicolumn{1}{l|}{} & U &
  KIS &
  0.4824 & 0.4963 &
  0.5007 &
  0.5025 &
  0.4991 &
  0.5034 &
  0.4999  \\
\multicolumn{1}{l|}{} & U & Pelt \cite{killick2012optimal,Wambui2015ThePO}      & 0.0952  & 0.0714                  & 0.0118     & 0.1223 &  0.1223     & 0.1430 & 0.1430 \\
\multicolumn{1}{l|}{} & U & BinSeg \cite{bai1997estimating,fryzlewicz2014wild}   &  0.0353 & 0.0353 & 0.0116     & 0.1295 & 0.1295       & 0.1422 & 0.1422   \\
\multicolumn{1}{l|}{} & U & Window \cite{truong2020selective}   & \textit{0.0109} & \textit{0.0109} & \textit{0.0109}     & \textit{0.0069} & \textit{0.0057}       & 0.0160 & \textit{0.0126}      \\
\multicolumn{1}{l|}{} & U & BottomUp \cite{keogh2001online}  &  0.0698 & 0.0581 & 0.1875     & \textbf{0.0000} & 0.1303       & 0.1430 & 0.1430        \\
\multicolumn{1}{l|}{} & U& IKSSW   & \textbf{0.0000} & \textbf{0.0000} & \textbf{0.0000}     & 0.0172 & \textbf{0.0000}       & \textit{0.0092} & \textbf{0.0000}      \\
\multicolumn{1}{l|}{\multirow{-6}{*}{\rotatebox{90}{Baselines}}} & U &
  IKSS-bdd \cite{dos2016fast} &
  \textbf{0.0000} &
  \textbf{0.0000} &
  \textbf{0.0000} &
  0.3330 &
  \textbf{0.0000} &
  0.1636 &
  \textbf{0.0000} \\
  \multicolumn{1}{l|}{} & U & KernelCPD \cite{arlot2019kernel} & 0.2024 & 0.0000 & 0.5316     & 0.4953 & 0.4953       & 0.4977 & 0.4977      \\
  \midrule
\multicolumn{1}{l|}{} & S & MD3 \cite{sethi2015don}  & 1.0000 & 1.0000 & 1.0000    & 0.0000 & 1.0000 & 1.0000 & 1.0000 \\
\multicolumn{1}{l|}{} & SS & MD3-EGM \cite{sethi2017reliable}  & 1.0000                     & 1.0000 & 1.0000     & \textbf{0.0000} & \textbf{0.0000}       & \textbf{0.0000} & \textbf{0.0000}   \\
\multicolumn{1}{l|}{} & SS & STUDD \cite{cerqueira2022studd}    & 1.0000                    & 1.0000 & 1.0000     & 0.6199 & 0.0091 & 0.3561 & 0.0899       \\
\multicolumn{1}{l|}{} & U & KLCPD \cite{DBLP:conf/iclr/ChangLYP19}    & \textbf{0.0000}                     & 0.0395 & $\times$ & 0.0330 & 0.3841       & 0.0292 & \textbf{0.0000}        \\
\multicolumn{1}{l|}{} & S & D3 \cite{gozuaccik2019unsupervised}       & 0.0112 & 0.0112                     & 0.0111 & 0.0243    & 0.1168 & 0.2786       & 0.1873           \\
\multicolumn{1}{l|}{} & U & NN-DVI \cite{liu2018accumulating}      & 0.2173 & 0.6250 & $\times$ & 0.0872 & 0.0720       & 0.1347 & 0.0884         \\
\multicolumn{1}{l|}{} & SS & ERICS \cite{haug2021learning}     & 0.9889 & 0.9889 & 0.9890 &     0.1324 &
  0.1187 &
  0.2591 &
  0.1872         \\
\multicolumn{1}{l|}{\multirow{-8}{*}{\rotatebox{90}{SoA}}} & U &
  CDLEEDS \cite{DBLP:conf/cikm/HaugBZK22}  &
  1.0000 &
  1.0000 &
  1.0000 &
  0.5771 &
  0.9967 &
  0.9967 &
  0.9967
  \\ \midrule
\multicolumn{1}{l}{} & U &
  DynAmo [us] &
  0.3667 &
  0.3667 & 
  0.2857 & 
  0.4578 &
  0.3128 &
  0.3893 &
  0.3927 \\ \bottomrule
\end{tabular}%
}
\end{table}

\begin{table}[!h]
\centering
\caption{Average FNR scores (over 30 runs) of SoA methods against DynAmo. Notice that $\times$ represents no convergence. A value is in bold if it is the lowest value on average; it is italic if it is the second best. S denotes supervised learning, SS semi-supervised learning, U unsupervised learning.}
\label{tab:fnr}
\resizebox{\linewidth}{!}{%
\begin{tabular}{@{}lclccccccc@{}}
\toprule
\multicolumn{3}{l}{}              & \multicolumn{3}{c}{Synthetic datasets}           & \multicolumn{4}{c}{Realistic datasets}                        \\ \cmidrule(lr){3-9}
\multicolumn{3}{l}{\multirow{-2}{*}{}} &
  PH &
  AS &
  VK &
  ELP1-D &
  ELP1-I &
  ELP2-D &
  ELP2-I \\ \midrule
\multicolumn{1}{l|}{} & U &
  KIS &
  0.5034 & 0.5004 &
  0.5046 &
  0.5027 &
  0.5014 &
  0.5024 &
  0.5044  \\
\multicolumn{1}{l|}{} & U & Pelt \cite{killick2012optimal,Wambui2015ThePO}      & 0.9114 & 0.9125                  & 0.9310     & 0.8703 & 0.8720       & 0.8584 & 0.8584 \\
\multicolumn{1}{l|}{} & U & BinSeg \cite{bai1997estimating,fryzlewicz2014wild}   &  0.9114 & 0.9250 & 0.9310     & 0.8567 & 0.8635       & 0.8567 & 0.8567   \\
\multicolumn{1}{l|}{} & U & Window \cite{truong2020selective}   & 0.9494 & 0.9875 & 0.9655     & 0.9932 & 0.9915       & 0.9881 & 0.9915      \\
\multicolumn{1}{l|}{} & U & BottomUp \cite{keogh2001online}  &  0.8861 & 0.9000 & 0.8966     & 0.8703 & 0.8720       & 0.8584 & 0.8584        \\
\multicolumn{1}{l|}{} & U& IKSSW   & 0.8734 & 0.8750 & 1.0000     & 0.9744 & 1.0000       & 0.9625 & 1.0000      \\
\multicolumn{1}{l|}{\multirow{-6}{*}{\rotatebox{90}{Baselines}}} & U &
  IKSS-bdd \cite{dos2016fast} &
  \textit{0.2125} &
  \textit{0.0247} &
  \textit{0.1525} &
  0.8703 &
  1.0000 &
  0.8481 &
  1.0000 \\
  \multicolumn{1}{l|}{} & U & KernelCPD \cite{arlot2019kernel} & 0.5316 & 0.6379 & 0.6000     & 0.5051 & 0.5051       & 0.5000 & 0.5000      \\
  \midrule
\multicolumn{1}{l|}{} & S & MD3 \cite{sethi2015don}  & 0.4595 & 0.1733 & 1.0000     & 1.0000 & 0.0630 & 0.0612 & 0.0072 \\
\multicolumn{1}{l|}{} & SS & MD3-EGM \cite{sethi2017reliable}  & 0.9857                     & 0.9859 & 0.9800     & 0.9982 & 0.9982       & 0.9982 & 0.9982   \\
\multicolumn{1}{l|}{} & SS & STUDD \cite{cerqueira2022studd}    & 0.9523                     & 1.0000 & 1.0000     & 0.1533 & 0.9838 & 0.7498 & 0.9234       \\
\multicolumn{1}{l|}{} & U & KLCPD \cite{DBLP:conf/iclr/ChangLYP19}    & 0.9163                     & 0.8148 & $\times$ & 0.9685 & 0.6138       & 0.9662 & 1.0000        \\
\multicolumn{1}{l|}{} & S & D3 \cite{gozuaccik2019unsupervised}       & 0.4430 & 0.4750                     & 0.3966     & \textbf{0.0683} & 0.3413     & 0.4590 & 0.6604          \\
\multicolumn{1}{l|}{} & U & NN-DVI \cite{liu2018accumulating}     & 0.3705 & 0.3208 & $\times$ & 0.6681 & 0.7116       & 0.8345 & 0.8507         \\
\multicolumn{1}{l|}{} & SS & ERICS \cite{haug2021learning}     & \textbf{0.0125} & \textit{0.0247} & \textbf{0.0169} &     0.7329 &
  0.6986 &
  0.6250 &
  0.6233         \\
\multicolumn{1}{l|}{\multirow{-8}{*}{\rotatebox{90}{SoA}}} & U &
  CDLEEDS \cite{DBLP:conf/cikm/HaugBZK22}  &
  0.6556 &
  0.6584 &
  0.5771 &
  0.1147 &
  \textbf{0.0000} &
  \textbf{0.0000} &
  \textbf{0.0000} 
  \\ \midrule
\multicolumn{1}{l}{} & U &
  DynAmo [us] &
  \textit{0.2125} &
  \textbf{0.0247} &
  \textit{0.1525} &
  \textit{0.0942} &
  \textit{0.0993} &
  \textit{0.0086} &
  \textit{0.2021} \\ \bottomrule
\end{tabular}%
}
\end{table}

\section{How does the window size affect the performance of SoA methods?}\label{sec:window_study}

\begin{figure*}
    \centering
    \includegraphics[width=\textwidth]{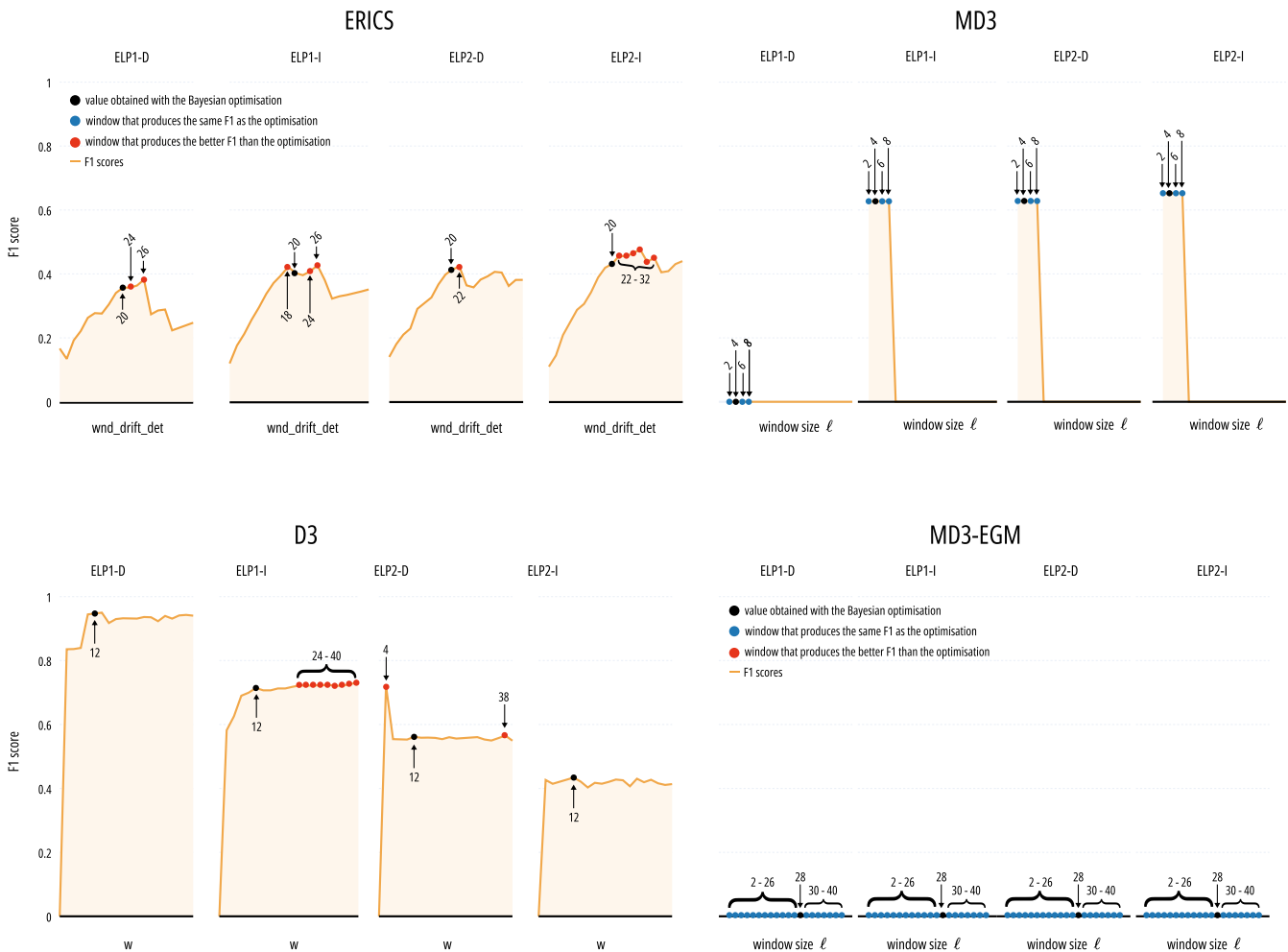}
    \caption{SoA's performances in terms of F1 scores when varying the window size hyperparameter for the systems not analyzed in Fig.  \ref{fig:limit_per_window}.}
    \label{fig:window_size_change}
\end{figure*}

This section shows how the drift detection window affects approaches that use the window size hyperparameter to signal drifts. For simplicity purposes, we show only how this hyperparameter affects the average F1 scores on the EL dataset. Figure \ref{fig:window_size_change} shows this effect for MD3, MD3-EGM, D3 and ERICS. Notice that we leave the other hyperparameters unvaried as specified in Table \ref{tab:soa_hyperparams}. Here, we set the window size in $[2,40]$ with a step of 2. We only illustrate the average F1 scores w.r.t. the variation of the window size for EL. We report with a black circle the average performances for all datasets reached according to the Bayesian optimisation (see Sec. \ref{sec:compared_methods}). Additionally, a blue circle illustrates similar performances as the ones reported from the optimisation with a different choice of window size. Red circles represent the choices of the window size, which produce better average results than those reported in Table 7 of the main material, leaving the other hyperparameters unchanged. 

The window size for MD3-EGM does not affect the performance at all. This means that no amount of additional information gives the underlying classifier how to separate the normal from the abnormal instances. The same happens for MD3 in ELP1-D since it never fires a drift signal (i.e., it has an average F1 of $0$). Generally, all methods have a non-decreasing trend regarding F1 scores when the window size increases. For ERICS, this happens until the trend reverses monotonicity. This is reasonable because, as discussed in Sec. \ref{sec:discussion} in the main material, drift detection approaches in real-world scenarios need to consider the promptness of the detection. Hence, a cutoff on the window size is an important aspect such that not much time is spent training/updating the models.

\section{Detecting Recurrent Anomalies}\label{sec:cyclic_anomaly_detection}

\begin{figure*}[!t]
    \centering
    \includegraphics[width=\textwidth]{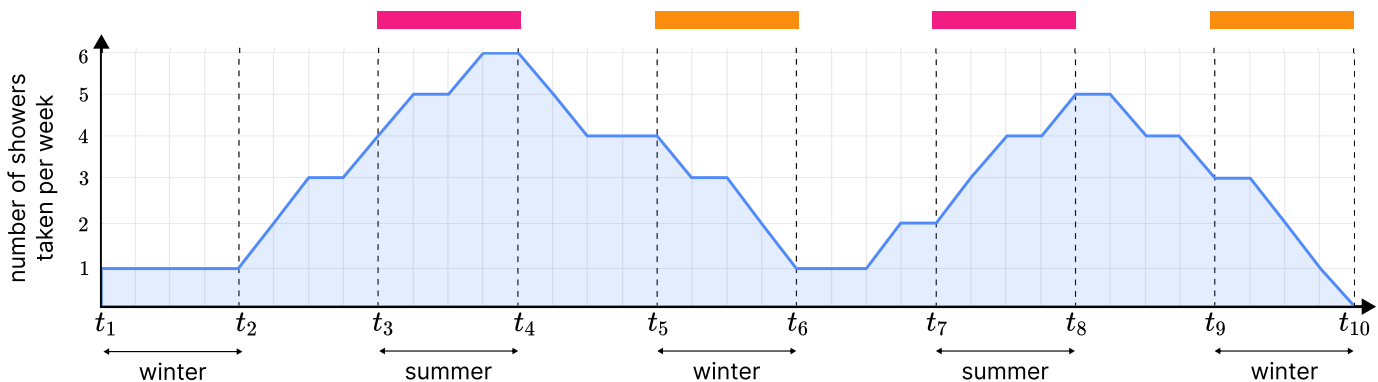}
    \caption{A toy example of recurrent anomalies corresponding to the shower/bathroom action in different seasons. For example, a patient might take, on average, more showers per week in the hot seasons rather than colder ones. The rectangles at the top of the series represent pairs of recurrent anomalies, e.g., a recurrent anomaly for $[t_3,t_4)$ is $[t_7,t_8)$ w.r.t. the initial reference window of $[t_1,t_2)$.}
    \label{fig:recurrent_anomalies}
\end{figure*}

DynAmo effectively detects recurrent anomalies through its two-window drift detection strategy. To illustrate this capability, consider Fig. \ref{fig:recurrent_anomalies}, which exhibits recurring behavioural shifts based on seasonality. For instance, patients may take more showers per week during hot seasons and fewer during colder ones. The analysis begins with a reference window representing the winter season at $[t_1,t_2)$. Monitoring commences by analyzing the evolution of hyperboxes within this time frame. The detection window, $[t_2,t_3)$, is subsequently labelled anomalous using the main material's divergence tests. This anomaly triggers a shift in the reference window, indicating a \textit{newly adopted normality}.

This process of the "new-normality-anomaly game" continues until the last time interval $[t_9,t_{10})$, where the entire series is labelled as anomalous based on the newly adopted behaviours. Such detection of recurring anomalies is crucial in healthcare, as addressing seasonality anomalies beforehand is essential for patient well-being. Although our toy example showcases a non-life-threatening anomaly, multiple other recurrent anomalies could cause permanent damage to patients \cite{karasmanoglou2023ecg,LIDDELL2016256}.

To enhance DynAmo's ability to reason on seasonality, we propose modifying its default behaviour to include a queue of past behaviours. When a new drift anomaly is detected, we can compare the distribution of new hyperboxes with those in the behaviour queue using statistical divergence tests like Kolmogorov-Smirnov. This process verifies whether the new anomaly has been observed previously, enabling us to identify recurrent anomalies within specific confidence intervals.

Naturally, the action to be taken upon detecting a recurrent anomaly depends on the specific deployment scenario and application. We leave the task of identifying recurrent anomalies and devising appropriate actions for future research, as it holds promise for further advancements in anomaly detection and healthcare practices.

\end{appendices}



%


\vfill


\end{document}